\pdfoutput=1
\documentclass[11pt]{article}

\usepackage[final]{acl}

\usepackage{times}
\usepackage{latexsym}

\usepackage[T1]{fontenc}
\usepackage[utf8]{inputenc}

\usepackage{microtype}

\usepackage{inconsolata}

\usepackage{graphicx}
\usepackage{booktabs}
\usepackage{amsmath}
\usepackage{amssymb}
\usepackage{subfig} 
\usepackage{enumitem}

\usepackage[hang,flushmargin]{footmisc}

\title{Time Course MechInterp: Analyzing the Evolution of Components and Knowledge in Large Language Models}

\author{
  Ahmad Dawar Hakimi \quad
  Ali Modarressi \quad
  Philipp Wicke \quad
  Hinrich Schütze \\
  Center for Information and Language Processing, LMU Munich \\
  Munich Center for Machine Learning \\
  \texttt{{adhakimi}@cis.lmu.de}
}

\newcounter{notecounter}
\newcommand{\enotesoff}{\long\gdef\enote##1##2{}}

\enotesoff

\begin{document}
\maketitle
\begin{abstract}
  Understanding how large language models (LLMs) acquire and store factual knowledge is crucial for enhancing their interpretability and reliability.
  In this work, we analyze the evolution of factual knowledge representation in the OLMo-7B model by tracking the roles of its attention heads and feed forward networks (FFNs) over the course of pre-training. 
  We classify these components into four roles: general, entity, relation-answer, and fact-answer specific and examine their stability and transitions. 
  Our results show that LLMs initially depend on broad, general-purpose components, which later specialize as training progresses. 
  Once the model reliably predicts answers, some components
  are repurposed, suggesting an adaptive learning
  process. Notably, attention heads
  display the
  highest turnover. We also present evidence that FFNs
  remain more stable throughout training.
 Furthermore, our probing experiments reveal that location-based relations converge to high accuracy earlier in training than name-based relations, highlighting how task complexity shapes acquisition dynamics. These insights offer a mechanistic view of knowledge formation in LLMs.
  We release the code and data.\footnote{\href{https://github.com/adhakimi/time-course-mechinterp}{https://github.com/adhakimi/time-course-mechinterp}}
\end{abstract}

\section{Introduction}

\begin{figure}[t]
    \centering
    \includegraphics[width=\columnwidth]{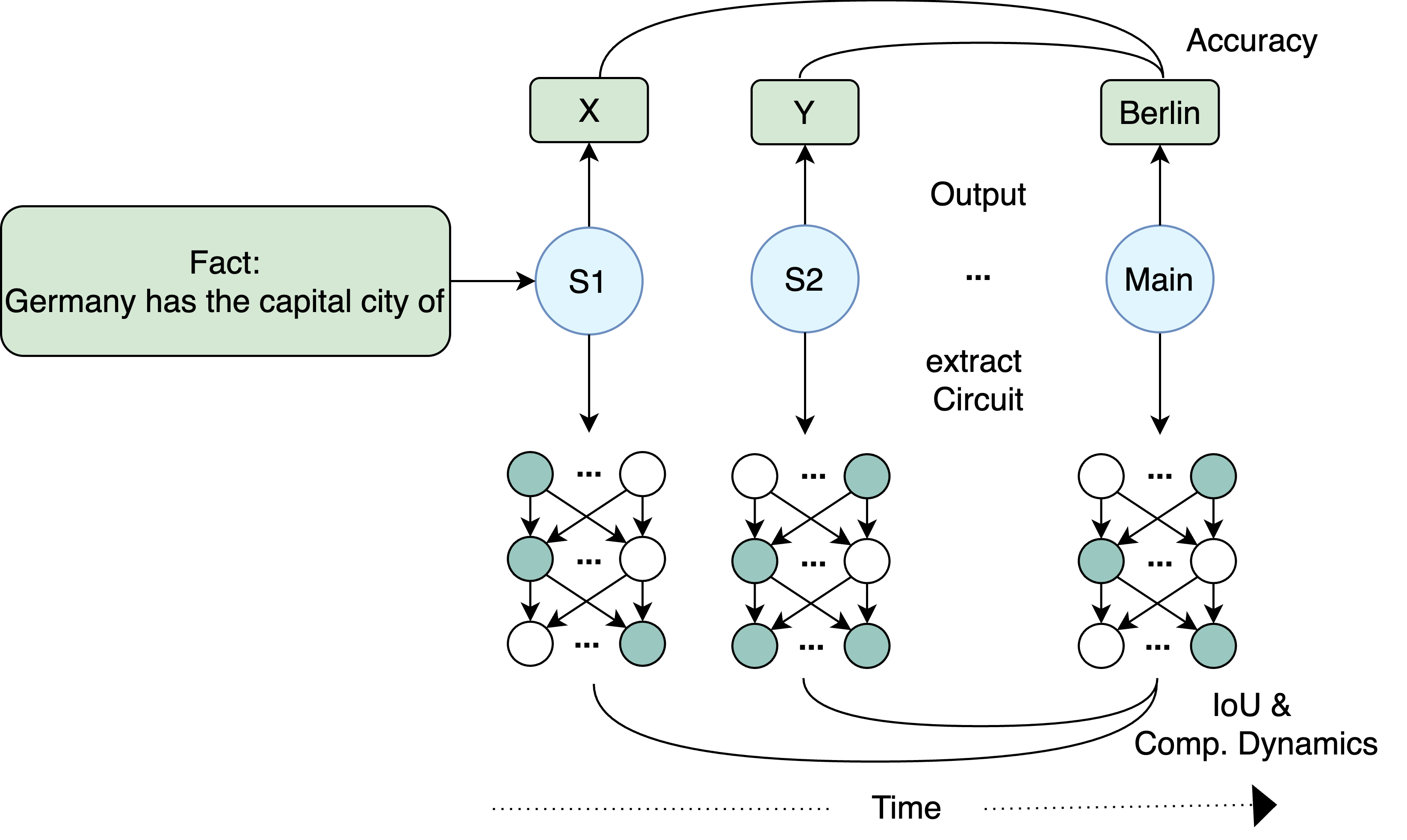}
    \caption{Factual Knowledge Probing. We trace how OLMo-7B processes factual knowledge across training snapshots by extracting Information Flow Routes and evaluating prediction accuracy. To analyze component dynamics, we measure the overlap of each component’s contributions with the fully trained model using Intersection over Union (IoU), and track how their assigned roles evolve over time.}
    \label{fig:fact_probing}
\end{figure}

Large language models (LLMs) are trained on vast datasets, including sources like Wikipedia, from which they acquire a wide range of factual knowledge. As a result, these models can provide informed answers when queried about facts. To uncover the mechanisms that enable such factual responses, mechanistic interpretability (MI) methods~\citep{olah:2020,elhage:2021} are employed. MI aims to reverse-engineer neural networks by translating their internal processes into human-understandable algorithms and concepts, and has made great progress in explaining how transformer-based LLMs process and store information.

A key approach within MI is \emph{circuit analysis}~\citep{olah:2020,elhage:2021,wang:2023}, which isolates minimal computational subgraphs comprising essential components like attention heads and FFNs that reproduce a model's behavior on a given task. Prior work on factual recall has focused on localizing knowledge within transformer parameters~\citep{meng:2022,geva:2021,geva:2022,geva:2023,hernandez:2024} and on behavioral analyses that trace the emergence of linguistic and reasoning capabilities during pretraining~\citep{rogers:2020,liu:2021,chiang:2020,chang:2024,xia:2023,hu:2023,biderman:2023}. While these studies have advanced our understanding of factual knowledge from both internal and external perspectives, they have not systematically examined how the components of a factual recall circuit evolve over training.

In this work, we bridge this gap by performing a time-course
mechanistic interpretability analysis of factual knowledge
representation in OLMo-7B~\citep{groeneveld:2024} by tracing
Information Flow Routes and classifying component roles over
40 training snapshots (see Figure \ref{fig:fact_probing}). Specifically, we investigate:

\begin{itemize}
    \setlength{\itemsep}{-1pt}
    \item Which components (attention heads, FFNs) contribute to solving factual knowledge tasks?
    \item How do these circuits for factual knowledge
      evolve over the course of model training?
\end{itemize}
To achieve this, we trace information flow routes
\cite{ferrando:2024} in OLMo-7B using interpretability
tools, analyze component dynamics, and classify components according to their roles within factual recall circuits across different training snapshots.
We make the following contributions:

\begin{itemize}
    \setlength{\itemsep}{-1pt}
    \item \textbf{Factual Knowledge Dataset:} We present a new probing dataset specifically designed for analyzing factual knowledge in LLMs, with prompts and examples carefully curated to minimize ambiguity.
    \item \textbf{Component Attribution for Factual Knowledge:} We analyze which components are responsible for processing factual knowledge at different training stages and the progression of factual knowledge acquisition over time. 
    \item \textbf{Temporal Evolution of Knowledge
      Representation:} We track how circuits responsible for
      factual knowledge stabilize and how the roles of their components  evolve over the course of training.
\end{itemize}

\section{Background}
\subsection{Circuit Analysis}
\label{sec:circuit-analysis}

A \emph{circuit} is defined as the minimal computational subgraph that faithfully reproduces a model's performance on a specific task~\citep{olah:2020,elhage:2021,wang:2023}. Circuits isolate key components such as attention heads and FFNs that drive predictions. Various techniques extract these circuits, including activation patching (which selectively corrupts activations to assess performance impact), attribution-based methods (e.g., edge attribution patching (EAP) and its integrated gradients variant, EAP-IG~\citep{hanna:2024,nanda:2023}), and gradient-based approaches like integrated gradients~\citep{sundararajan:2017}. However, these methods often become computationally prohibitive for large models or when evaluating multiple snapshots due to their complexity and memory demands.

\subsection{Information Flow Routes}
\label{sec:information-flow-routes}

To overcome these limitations, we leverage Information Flow Routes (IFRs)~\citep{ferrando:2024}. IFRs conceptualize the model as a computational graph and recursively trace pathways from the output token back through the network. At each step, only nodes and edges with contributions exceeding a threshold $\theta$ are retained, ensuring that only paths significantly impacting the final prediction are included. The importance of each edge is quantified using a modified ALTI (Aggregation of Layer-Wise Token-to-Token Interactions) score~\citep{ferrando:2022}. Compared to traditional circuit-finding methods, IFRs are more scalable, require minimal prompt design, and are well-suited for large models like OLMo-7B across multiple training snapshots. Furthermore, IFRs sidestep challenges posed by self-repair mechanisms in LLMs~\citep{mcgrath:2023,rushing:2024}, making them a robust tool for circuit analysis.

\section{Factual Knowledge Probing over Time}
In this section, we describe our approach to probing factual knowledge (Fig.~\ref{fig:fact_probing}). We first introduce our dataset (Sec.~\ref{sec:dataset}), then detail the OLMo-7B training snapshots used (Sec.~\ref{sec:models}), and finally assess snapshot performance via accuracy (Sec.~\ref{sec:accuracy}).

\paragraph{Key Terms.}
A \textbf{fact} is defined as a subject-relation-object triple (e.g., (Canada, has\_capital, Ottawa)), where \textbf{has\_capital} is the \textbf{relation} representing the pairing of a country with its capital.

\paragraph{Token Positions.}
In our experiments, facts appear in sentences such as “Canada has the capital city of Ottawa.” We distinguish three sets of subtoken positions: (i) \textbf{SUBJECT} for the subject (e.g., “Canada”); (ii) \textbf{END} for the subtoken immediately before the answer (e.g., “of” in “has the capital city of”); and (iii) \textbf{ANSWER} for the tokens forming the answer, beginning with the token following \textbf{END}.

\subsection{Dataset}\label{sec:dataset}

\begin{table*}[h]
  \centering
  \footnotesize
  \renewcommand{\arraystretch}{0.9}
  \setlength{\tabcolsep}{2.5pt}
  \resizebox{\textwidth}{!}{
  \begin{tabular}{@{}llcl@{}}
  \toprule
  \multicolumn{4}{c}{\textbf{Location-based Relations (LOC)}} \\[3pt]
  \midrule
    \textbf{Relation}         &   \textbf{Prompt Template}                                      & \textbf{\# Facts}  & \textbf{Example Subject}    \\
  \midrule
    CITY\_IN\_COUNTRY         &   \{\} is part of the country of                                &   14    & Rio de Janeiro, Buenos Aires                 \\
    COMPANY\_HQ               &   The headquarters of \{\} are in the city of                   &   20    & Zillow, Bayrischer Rundfunk                   \\
    COUNTRY\_CAPITAL\_CITY    &   \{\} has the capital city of                                  &   19    & Canada, Nigeria                               \\
    FOOD\_FROM\_COUNTRY       &   \{\} is from the country of                                   &   17    & Sushi, Ceviche                                \\
    OFFICIAL\_LANGUAGE        &   In \{\}, the official language is                             &   14    & France, Egypt                                 \\
    PLAYS\_SPORT              &   \{\} plays professionally in the sport of                     &   12    & Kobe Bryant, Roger Federer                    \\
    SIGHTS\_IN\_CITY          &   \{\} is a landmark in the city of                             &   17    & The Eiffel Tower, The Space Needle            \\
  \midrule
  \multicolumn{4}{c}{\textbf{Name-based Relations (NAME)}} \\[3pt]
  \midrule
    \textbf{Relation}         &   \textbf{Prompt Template}                                      & \textbf{\# Facts}  & \textbf{Example Subject}    \\
  \midrule
    BOOKS\_WRITTEN            &   The Book \{\} was written by the author with the name of       &   13    & The Hunger Games, Life of Pi                  \\
    COMPANY\_CEO              &   Who is the CEO of \{\}? Their name is                           &   17    & Ubisoft, Pinterest                            \\
    MOVIE\_DIRECTED           &   The Movie \{\} was directed by the director with the name of    &   17    & The Godfather, Forrest Gump                    \\
  \bottomrule
  \end{tabular}
  }
  \caption{Overview of the Factual Knowledge dataset, grouped by relation type.}
  \label{tab:dataset}
\end{table*}

We develop a dataset designed to probe the factual knowledge 
encoded in a given LLM. 
See Table \ref{tab:dataset}.
To minimize syntactic ambiguity, 
we avoid templates that may lead to multiple valid answers; e.g., 
for the prompt \emph{"The Eiffel Tower is located in"}, both
\emph{Paris} and \emph{France} are correct.
Similarly, we avoid cases involving
regional variations in terminology (e.g., \emph{soccer}
vs. \emph{football}) and eliminate instances where the
answer is already contained in the subject (e.g.,
\emph{"The Leaning Tower of Pisa is a landmark in the city
of Pisa."}). Our focus is on categorical facts associated
with well-defined relation types, specifically
\textbf{Location-Based Relations} (LOC) and
\textbf{Name-Based Relations} (NAME). Table~\ref{tab:dataset} provides an overview of these relations, along with the prompt templates, number of facts, and example subjects for each relation type. Although manually
curated, our dataset is inspired by existing resources such
as \textbf{LRE}~\cite{hernandez:2024},
\textbf{CounterFact}~\cite{meng:2022}, and 
\textbf{ParaRel}~\cite{elazar:2021}, as well as
\textbf{Summing Up The Facts}~\cite{chughtai:2024}. We extended these resources by integrating  relations such as \texttt{BOOKS\_WRITTEN} and \texttt{MOVIE\_DIRECTED} using data from Goodreads and IMDb's Top Favorites list. To ensure reliability and eliminate potential confounds in our
analysis, we implement a multi-step validation pipeline to
rigorously evaluate both the prompts and the facts (see
Appendix~\ref{app:dataset-construction-pipeline}).

\subsection{Models}\label{sec:models}

We study the evolution of factual knowledge using the \textbf{OLMo-7B} model~\cite{groeneveld:2024},\footnote{\url{https://huggingface.co/allenai/OLMo-7B-0424-hf/tree/main}} a flagship open-source LLM with 32 layers (each with 32 attention heads) pretrained on over 2.5 trillion tokens.
During training, checkpoints were saved every 500 steps (2B tokens per interval) from initialization up to \texttt{step161000-tokens675B}, and then at 1000-step intervals until the final checkpoint, \texttt{step651581-tokens2731B}. For our analysis, we select 40 snapshots spanning from \texttt{step5000-tokens20B} to \texttt{step200000-tokens838B} (in 5000-step increments), along with the fully trained main model.\footnote{Due to an issue with \texttt{step115000}, we use \texttt{step115500-tokens462B} instead.} We denote these snapshots as SX-YB, where X represents the training step (in multiples of 5000) and Y the token count in billions.

\subsection{Accuracy}
\label{sec:accuracy}

Since we are interested in the time course of factual knowledge during training, we first 
establish
how the acquisition of knowledge evolves as measured by top-1 and top-10 accuracy on the first 
token of ANSWER. We group our relations into two groups: NAME (the answer is the name of a person) and LOC (the answer is a location).
See Appendix~\ref{app:accuracy-plots-per-relation} for per-relation graphs.

\begin{figure*}[t]
    \centering
    \includegraphics[scale=0.3]{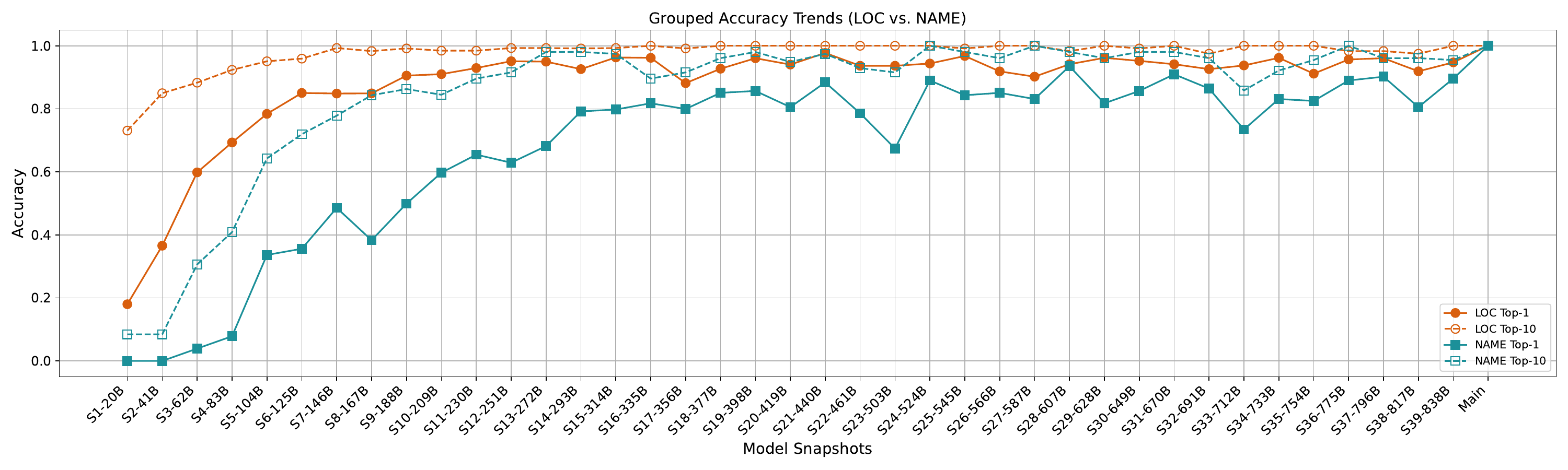}
    \caption{Top-1 and top-10 accuracy for location-based (LOC, orange) and name-based (NAME, teal) factual relations across 40 OLMo-7B snapshots and the fully trained model. Solid lines show top-1 accuracy (circles for LOC; squares for NAME), while dashed lines show top-10 accuracy with hollow markers.} 
    \label{fig:accuracy-values}
\end{figure*}

Figure~\ref{fig:accuracy-values} shows that LOC relations
converge faster than NAME relations: A top-1 accuracy of 0.8
is first reached at S5 for LOC and at S14 for NAME.  LOC
also has less top-1 volatility than NAME.  top-10 accuracy
for NAME is also lower than for LOC, but top-10 values are much
higher, starting at about S13.  This indicates that fairly
early on, the correct NAME is in the pool of candidates that
the model has identified as relevant and that the remaining
problem of knowledge acquisition is then correct ranking.
The likely reason for these differences between NAME and LOC
is that there are many more prominent person names than
prominent locations in the model's training data (see Appendix~\ref{app:counts}), making it more challenging to learn the correct answer for a person than for a location.

\section{How do Components Evolve?}
We now examine OLMo-7B's internal mechanics during
pretraining. Using IFR, we trace the full circuit behind
each predicted token to identify the contributing
components, i.e., 
attention
heads and FFNs.
We classify these components 
based on their roles in the circuit,
distinguishing generalized components that contribute
broadly from specialized ones with more focused
functions. By tracking how these roles evolve during
pretraining, we gain deeper insights into the model’s
learning dynamics for factual knowledge.

\subsection{Model Component Roles}

We take a systematic approach to defining model component
roles based on their contributions within token
circuits. These roles are determined by the types of tokens
a component influences and the scope of its contribution. A
component may contribute to all tokens or to a subset,
to only one fact or multiple facts etc.
We now define a
structured classification schema that captures the
functional behavior of each component.

For each snapshot $s$,
relation $r$,
fact $f$
and subtoken position $t$,
we use IFR to compute the circuit that produced the output
subtoken at that $t$.
We set
$c_{srft}=1$ if component $c$ is part of the circuit, 0
otherwise.

We then define
$c_{srf}(T)= \frac{1}{|T(f)|} \sum_{t \in T(f)} c_{srft}$, i.e., $c_{srf}$
is
the activation of $c$ averaged over the subtoken positions $T(f)$.
Given the sentence
corresponding to fact $f$,
$T(f)$ is a  subset of its subtokens.
We now define different roles of a component by defining
$T(f)$ differently, e.g., containing only the ANSWER
subtokens or all subtokens.

\subsubsection{General role}
For the general role, we
use the subtoken selector $T_g(f)$.
$T_g(f)$ is the set of all subtokens of the sentence
(except for
the final period).

We define the general activation score of a component $c$ for
snapshot $s$ as:
\[
c^g_{s}= \frac{\sum_{r \in R} \sum_{f \in r} c_{srf}(T_g)}{\sum_{r \in R} \sum_{f \in r} 1}
\]
That is, $c^g_s$ is the activation of $c$ for snapshot
$s$, microaveraged over facts. $R$ is the set of relations.

We classify a component $c$ as having a \textbf{general role} for snapshot $s$ if
$c^g_s>\theta$, where we set $\theta = 0.1$.

\subsubsection{Entity role}

For the entity role, we
use the subtoken selector $T_e(f)$.
$T_e(f)$ is the set of all subtokens of
SUBJECT and ANSWER.\footnote{For the SUBJECT, there is no helpful context for the prediction of its
first subtoken, e.g., 
for
``France''
in 
``France has the capital \ldots''.
We therefore shift the subtokens considered to the right by 1 for SUBJECTS.}

We define the entity activation score of a component $c$ for
snapshot $s$ as:
\[
c^e_{s}= \frac{\sum_{r \in R} \sum_{f \in r} c_{srf}(T_e)}{\sum_{r \in R} \sum_{f \in r} 1}
\]
That is, $c^e_s$ is the subject and answer activation of $c$ for snapshot
$c$, microaveraged over facts.

We classify a component $c$ as having an \textbf{entity role} for snapshot $s$ if
$c^e_s>\theta$ where $\theta = 0.1$.

\subsubsection{Relation-answer specific role}
For the relation role, we
use the subtoken selector $T_a(f)$.
$T_a(f)$ selects the subtokens
of the ANSWER.
We then define the relation-answer activation score of a component $c$ for
snapshot $s$ and relation $r$ as:
\[
c^r_{s}= \frac{\sum_{f \in r} c_{srf}(T_a)}{\sum_{f \in r} 1}
\]
That is, $c^r_s$ is the answer activation of $c$ for snapshot
$s$ and relation $r$, averaged over facts.

We classify a component $c$ as having a \textbf{relation-answer role}
if
$c^r_s>\theta$ where  $\theta = 0.1$.

\subsubsection{Fact-answer specific role}
For a fact $f$ belonging to relation $r$,
we set
$c^f_{s} =  c^f_{srf}(T_a)$.

We classify a component $c$ as having a \textbf{fact-answer role}
if
$c^f_s>\theta$ where $\theta = 0.1$.

\subsubsection{Proper Components}
Let
${\cal J}_g$,
${\cal J}_e$,
${\cal J}_r$ and
${\cal J}_f$
be the sets of components that assume the general, entity, relation-answer, and fact-answer roles, respectively, as defined above.

We define the set of proper entity components ${\cal H}_e$ as those components that assume an entity role but not a general role:
\[{\cal H}_e = {\cal J}_e - {\cal J}_g\]

We define the set of proper relation-answer components ${\cal
  H}_r$ as those components that assume a relation-answer role but not an entity or general role:
\[{\cal H}_r =  {\cal J}_r - {\cal J}_e - {\cal J}_g\]

We define the set of proper fact-answer components ${\cal
  H}_f$ as those components that assume a fact-answer role but not a general, entity, or relation-answer role:
\[{\cal H}_f =  {\cal J}_f - {\cal J}_r - {\cal J}_e - {\cal J}_g\]

We set:
\[{\cal H}_g =  {\cal J}_g\]
because for the general role, there is no change from the
original set to the proper set.

Finally, 
in addition to the sets of components
\(\mathcal{H}_g\), \(\mathcal{H}_e\), \(\mathcal{H}_r\), and
\(\mathcal{H}_f\), we define the set of deactivated
components \(\mathcal{H}_d\)
as the complement of the union of the four other roles $g$, $e$,
$r$, $f$: 
\(\mathcal{H}_d
= {\mathcal C} (
\mathcal{H}_g \cup
\mathcal{H}_e \cup
\mathcal{H}_r \cup
\mathcal{H}_f)\).

\subsection{Analysis of Component Dynamics}
After classifying components into five distinct roles
(general, entity, relation-answer,  answer-specific, and deactivated), we
analyze how these
components evolve during pretraining,
quantifying both static and dynamic aspects of the roles.
We now describe our methodology including the measures we use.

\paragraph{Consistency and Count Metrics:}  
To quantify stability, we measure the consistency of each
role over time by calculating the Jaccard Similarity
(Intersection over Union, or IoU) between the set of
components with a particular role at a given snapshot and
the corresponding set in the final model. For example, for general components, the IoU is defined as:
\[
\text{IoU}({\cal H}_{g}) = \frac{{\cal H}_{gs} \cap {\cal H}_{gmain}}{{\cal H}_{gs} \cup {\cal H}_{gmain}},
\]
where \({\cal H}_{gs}\) represents the set of entity components in the current snapshot, and \({\cal H}_{gmain}\) is the corresponding set in the final model.

\paragraph{Role Switch Dynamics:}  
We also track how components change roles over time: whether
they activate, deactivate, or switch functions. By computing
the accumulated switch counts across selected snapshots
(S1, S10, S20, S40, and the
main model), we capture the dynamics of these transitions,
such as deactivated components reactivating in specialized
roles or components switching between different roles.

\paragraph{Markov Chain Modeling of Transitions:}  
To further characterize role dynamics, we model transitions
using a Markov chain.
The transition probability from state
\(\mathcal{H}_\alpha\) to state \(\mathcal{H}_\beta\) is
given by:
\[
P\bigl(\mathcal{H}_\alpha \to \mathcal{H}_\beta\bigr) = \frac{N\bigl(\mathcal{H}_\alpha \to \mathcal{H}_\beta\bigr)}{\sum_{\gamma \in \{g,e, r, f, d\}} N\bigl(\mathcal{H}_\alpha \to \mathcal{H}_\gamma\bigr)},
\]
where \(N(\mathcal{H}_\alpha \to \mathcal{H}_\beta)\) is the
number of observed transitions from one state to the
following state.

\subsection{Temporal Consistency and Role Dynamics of Attention Heads}

\begin{figure*}[t]
    \centering
    \includegraphics[width=\textwidth]{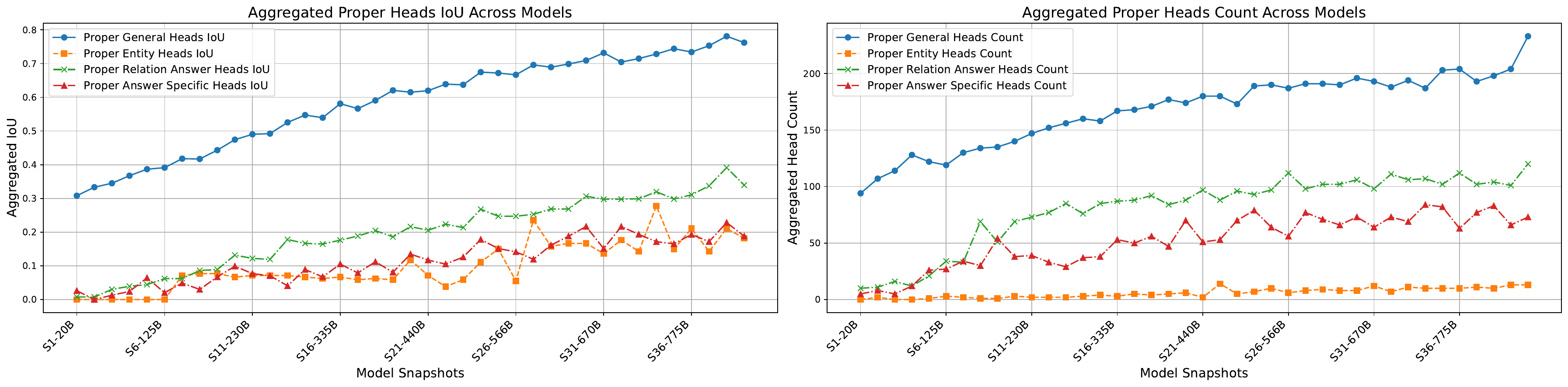}
    \caption{Aggregated Head Count and IoU Across OLMo-7B Snapshots
    Left: IoU values comparing each snapshot to main.
    Right: Counts for general, entity, relation-answer, and answer-specific categories over snapshots.
    }
    \label{fig:aggregated_combined}
\end{figure*}

Our analysis reveals key trends in the evolution of attention heads. Since differences between LOC and NAME relations are marginal (see Appendix~\ref{app:relation-level-counts-iou}), we combine them for the subsequent analysis.

Using the IoU metric and component counts, we observe that the number of active attention heads increases steadily over the course of training. For instance, the counts for general heads rise from 94 to 233, for relation-answer heads from 8 to 78, and for answer-specific heads from 11 to 99. Overall, the total number of active heads grows from 113 to 423 rising from approximately 11\% to 41\% of all heads while nearly 60\% remain deactivated. Figure~\ref{fig:aggregated_combined} illustrates these dynamics, with the IoU metric confirming that general heads maintain a high consistency with the final model throughout training.

The evolution of attention head roles suggests a
hierarchical learning process. Early in training, the model
primarily relies on general-purpose heads that generate
broad, context-independent representations. As training
progresses, specialized heads emerge to support more precise
fact retrieval. Notably, answer-specific heads demonstrate
the highest turnover 
(even the IOU between the final model and the checkpoint
just before the final
model is just 0.2), indicating frequent role changes and dynamic reallocation of resources. Furthermore, our observations indicate that tasks involving complex, name-based relations require longer training periods (Fig.~\ref{fig:accuracy-values}) and exhibit more frequent role transitions compared to simpler, location-based tasks (Fig.~\ref{fig:switches-name_app}).

\paragraph{Dynamic Specialization and Generalization of Attention Heads}

\begin{figure}[t]
    \centering
    \includegraphics[width=\columnwidth]{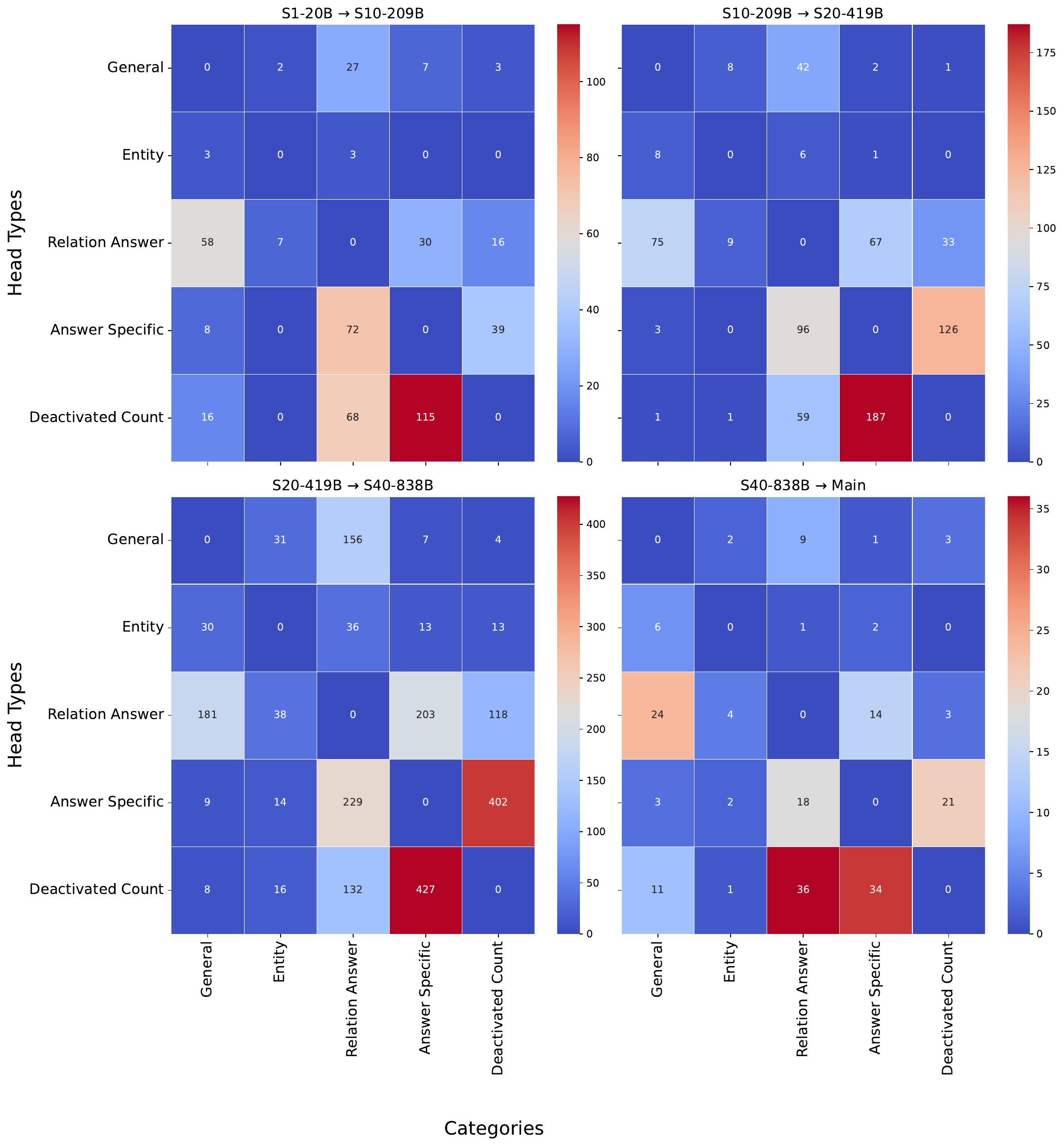}
    \caption{Accumulated Attention Head Switches Across Training Stages. Heatmaps showing the total number of transitions between the four types of  heads and the deactivated state at key training snapshots (S1, S10, S20, S40, and Main).}
    \label{fig:accumulated_switch_count}
\end{figure}

\begin{figure}[t]
    \centering
    \includegraphics[width=\columnwidth]{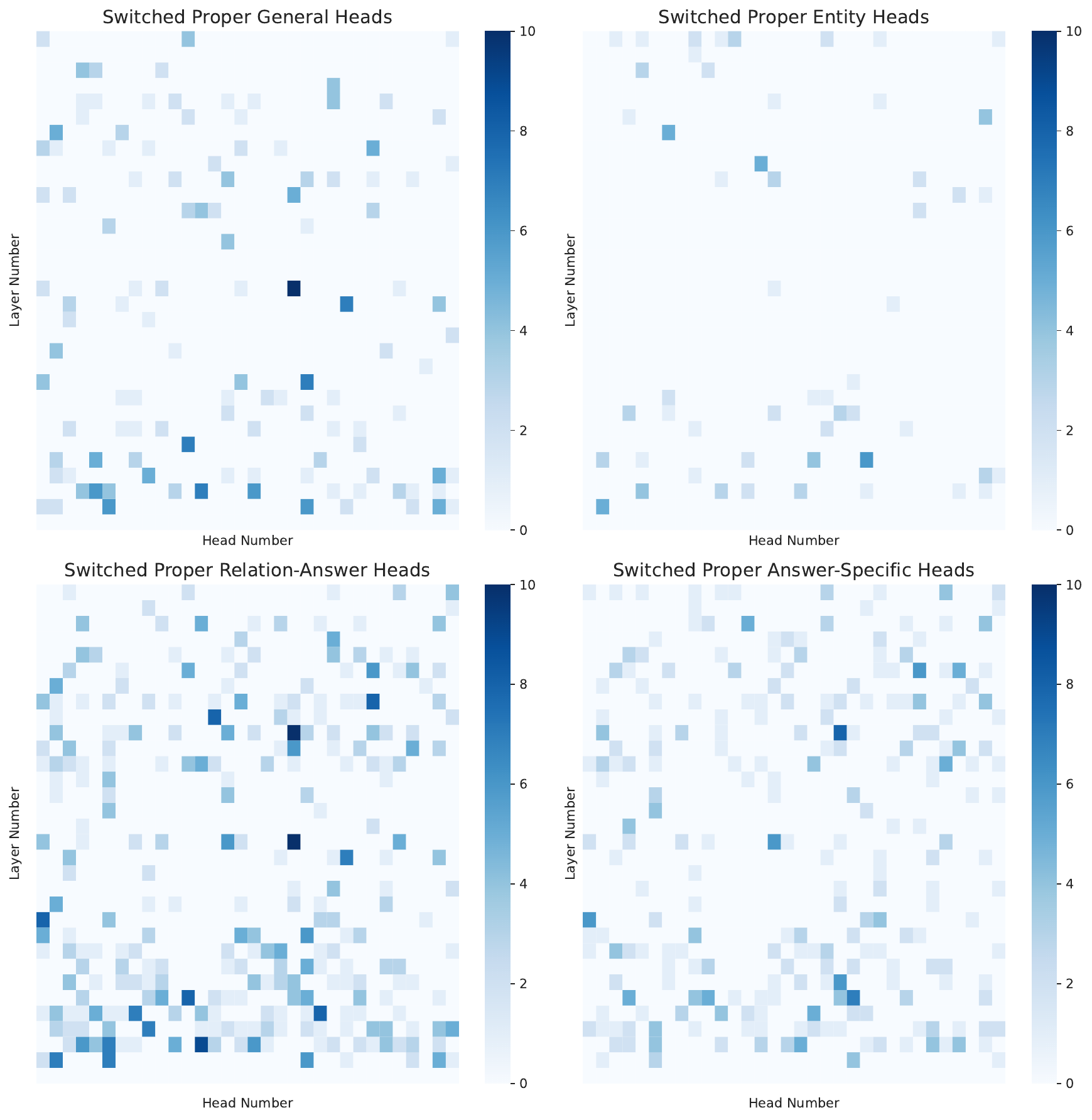}
    \caption{Attention Head Role Transitions. Per-layer heatmaps
      showing the frequency
that a head from one of the
four roles  general, entity, relation-answer, and
answer-specific switches to a different role.}
    \label{fig:switches}
\end{figure}

Fig.~\ref{fig:accumulated_switch_count}) shows that attention heads frequently
transition from deactivated to specialized roles especially
to answer-specific roles (however, they also frequently
shift back from specialized to deactivated). In contrast,
general heads are more stable or shift to relation-answer
roles. Figure \ref{fig:switches} illustrates that early and
late layers undergo frequent switching, whereas middle
layers (10–18) remain comparatively stable. Notably, in these mid-layers, NAME-based tasks both engage more heads in active roles and exhibit higher switching among those roles compared to LOC-based tasks, reflecting the greater complexity of person-name relations and the additional head capacity required to process them. See
Appendix~\ref{app:attention-head-switches} for details.

\begin{figure}[t]
    \centering
    \includegraphics[width=0.99\columnwidth]{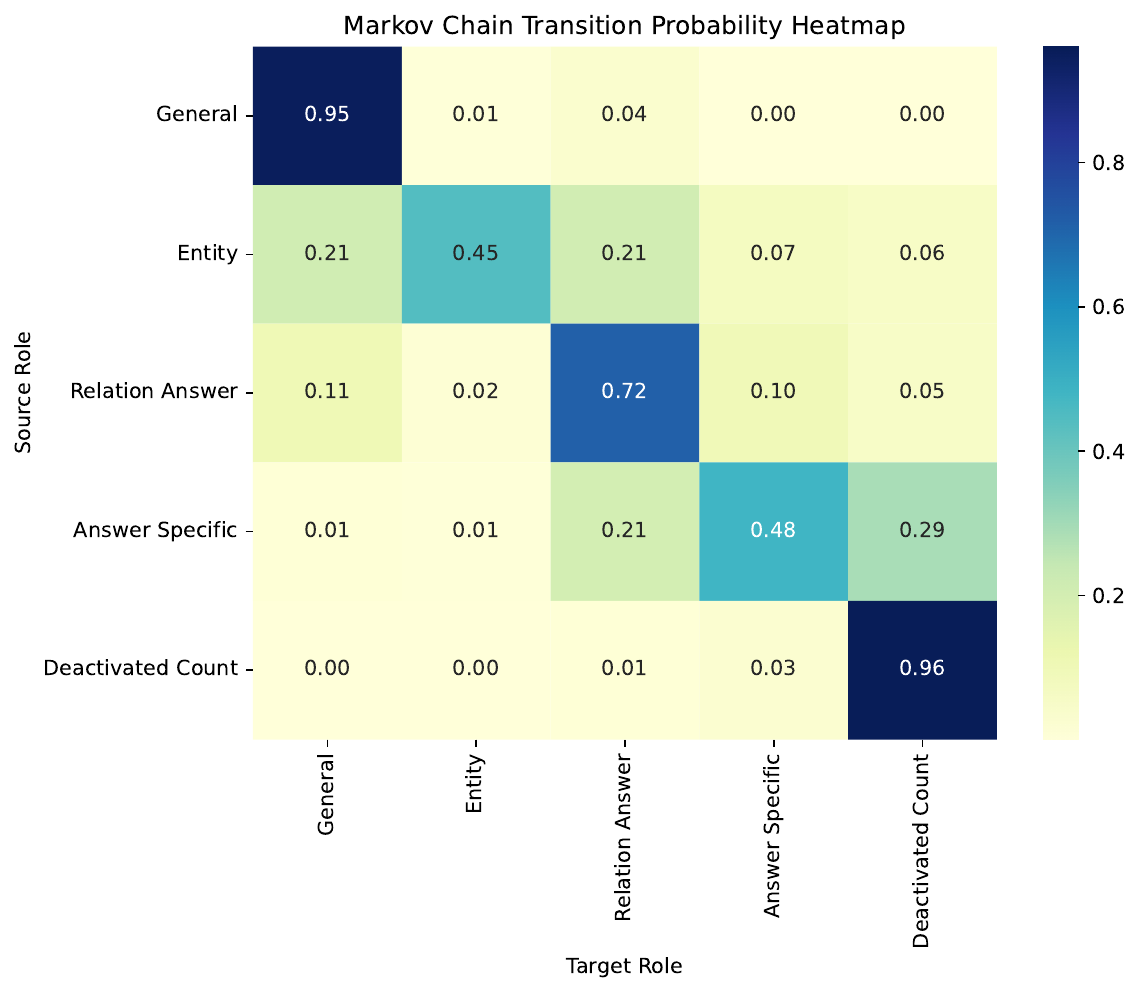}
    \caption{Markov Chain Transition Probability Heatmap.
    Heatmap showing the transition probabilities between
    different attention head roles across model
    snapshots. Each cell represents the probability of a
    head transitioning from a source role (rows) in snapshot
    $i$ to a target
    role (columns) in snapshot  $i+1$.}
    \label{fig:transition_probability_heatmap_all_heads}
\end{figure}

Our Markov chain modeling (see
Fig.~\ref{fig:transition_probability_heatmap_all_heads})
further quantifies these dynamics: specialized heads tend to
transition toward more general roles.
Although individual specialized
heads often shift into general roles, the overall count of
specialized heads increases over time because the rate at
which new specialized heads emerge exceeds the rate at which
they move to other roles.

Overall, our findings indicate that attention heads first converge on broad, general-purpose roles and only later diversify into specialized functions.%, illustrating a progression from stability toward targeted flexibility.

\subsection{FFNs over Time}
Analogous to our attention head classification, we assign FFNs to five roles (general, entity, relation-answer answer-specific, and deactivated), though with a higher activation threshold (\(\theta = 0.90\)) as suggested in \cite{ferrando:2022}. Unlike the 1024 attention heads, the model uses only 32 FFNs (one per layer), and all actively contribute to answer generation.

\begin{figure*}[t]
    \centering
    \includegraphics[width=\textwidth]{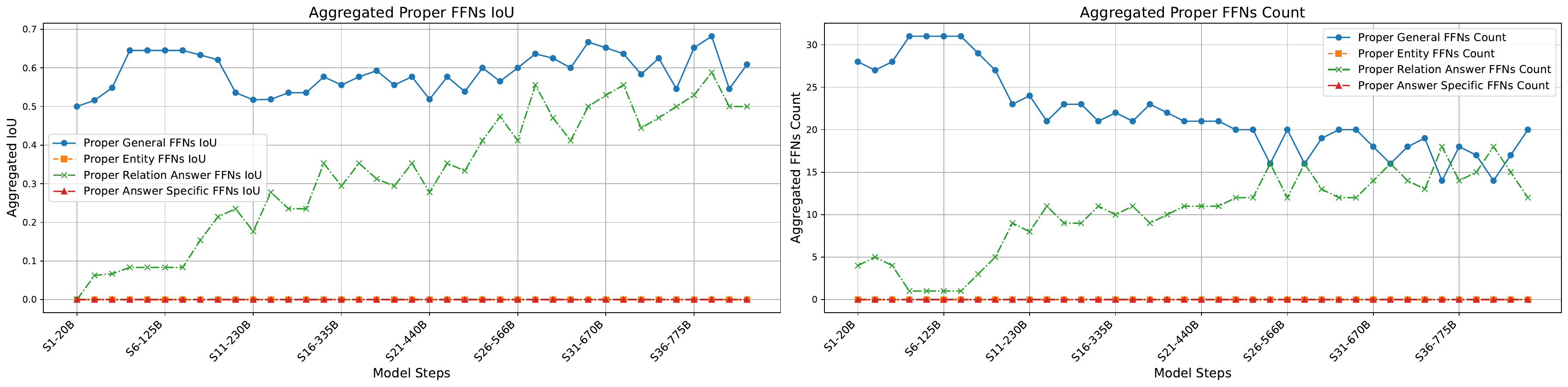}
    \caption{Aggregated FFN Count and IoU Across OLMo-7B Snapshots.
    Left: IoU values comparing each snapshot to main.
    Right: Counts for general, entity, relation-answer, and answer-specific categories over snapshots.}
    \label{fig:aggregated_ffn_combined}
\end{figure*}

\paragraph{Steady Backbone: Consistency and Activation Trends}
Figure~\ref{fig:aggregated_ffn_combined} shows that early
on, most FFNs serve as general components, with only a few
operating in relation-answer or answer-specific
roles. Around stages S7–S8, when accuracy exceeds 80\%, many
general FFNs shift to relation-answer roles.

\paragraph{Oscillatory Dynamics in Role Allocation}
Although the majority of FFNs remain general, we observe occasional role oscillations. For easier LOC relations, answer-specific FFNs exhibit minimal switching, whereas for the more challenging NAME relations, a small number of FFNs gradually transition into answer-specific roles before reverting to general roles in subsequent pretraining steps. The total switch count and transition probability analyses (see Appendix~\ref{app:ffn-role-transitions}) show that FFNs rarely stay specialized; in most cases, they revert back to the general role rather than maintain a specialized function.

\paragraph{FFNs as General Processing Components}
In contrast to the dynamic specialization observed in
attention heads, FFN behavior looks more stable
in Figure 7.
This is an apparent 
contrast with the highly dynamic behavior of attention
heads.
We could interpret this as
the FFNs acting as a
steady processing backbone that incrementally refines the
representations produced by attention.
However, it is important to consider two caveats.
First,
our high fixed
activation threshold may accentuate the apparent
generality of FFNs. Second, the more components are lumped
together into larger units, the more likely it is that our
analysis will classify them as general. FFNs consist of
many individual neurons and therefore our methodology may be
prone to classify them as general.

\section{Related Work}
This section reviews prior work on mechanistic
interpretability, model behavior evolution, and how
transformers store and retrieve factual knowledge. While
past research has deepened our understanding of fully
trained models, less focus has been given to how these
mechanisms evolve during training -- a gap this work addresses.

\subsection{Mechanistic Interpretability} 
Mechanistic Interpretability aims to reverse-engineer neural networks to uncover circuits driving model behavior. Early work~\citep{elhage:2021,olah:2020} focused on vision models and has since extended to transformer language models~\citep{meng:2022,wang:2023,hanna:2023,varma:2023,merullo:2024,lieberum:2023,tigges:2023,mondorf:2024,tigges:2024}. Research has characterized attention heads~\citep{olsson:2022,chen:2024,singh:2024,gould:2024,mcdougall:2023,chughtai:2024,elhelo:2024,ortu:2024} and FFNs~\citep{geva:2021,meng:2022,bricken:2023,neo:2024,tian:2024}.

\subsection{Interpretability Over Time}

Behavioral studies have tracked the emergence of linguistic and reasoning capabilities during pretraining~\citep{rogers:2020,liu:2021,chiang:2020,muller-eberstein:2023,xia:2023,chang:2023,hu:2023,biderman:2023}. Yet, they offer limited insight into internal circuit evolution. Recent work on smaller models shows that internal circuits can change abruptly even when overall behavior is stable~\citep{nanda:2023,olsson:2022,chen:2024} and mechanistic studies have begun tracking circuit evolution~\citep{tigges:2024}.

\subsection{Mechanisms of Knowledge Storage in Transformers}
Studies have shown that transformers store factual knowledge in (subject, relation, attribute) tuples. Causal interventions reveal that early-to-middle FFNs enrich subject representations, while attention heads pass relation information and later layers extract attributes~\citep{meng:2022,geva:2023}. Complementary work demonstrates that these representations can be decoded to recover facts~\citep{hernandez:2024,chughtai:2024}. Other research highlights the balance between in-context and memorized recall~\citep{yu:2024,variengien:2023} and the distributed nature of knowledge retrieval~\citep{haviv:2023,stoehr:2024,chuang:2024}. 

While previous work has focused on static models, we track the evolution of these mechanisms during training, offering a dynamic view of factual knowledge development in LLMs.

\section{Discussion \& Conclusion}
Our time-course mechanistic interpretability study of OLMo-7B uncovers several core patterns in how factual knowledge circuits form and evolve during pretraining:
\begin{enumerate}[topsep=5pt, itemsep=-2pt]
    \item \textbf{Task Complexity Influences Training Dynamics:} Location-based relations are acquired more rapidly and stably than name-based relations, which require more specialized components.
    \item \textbf{Hierarchical Learning Process:} Early training is dominated by stable, general attention heads that lay the groundwork for subsequent specialization.
    \item \textbf{Adaptive vs. Stable Components:} Our analysis suggests that certain attention heads may be repurposed dynamically particularly those associated with answer-specific roles while FFNs may exhibit a more stable behavior. These observations hint at a possible complementary dynamic between adaptable attention mechanisms and stable processing components. %However, we note that further investigation is required to fully substantiate and generalize these findings.
    \item \textbf{Evolving Specialization:} In the later stages of training, the model stabilizes into a configuration dominated by general heads. While many components retain their roles, we also observe that a substantial number of previously deactivated heads reactivate and adopt specialized functions, highlighting the continued plasticity of the model even in later training phases.
\end{enumerate}

\noindent These insights advance our understanding of mechanistic interpretability by showing that dynamic specialization in attention heads supported by consistent FFN refinement underpins effective factual knowledge retrieval. Future work may explore neuron-level dynamics, assess redundancy among head roles, and examine scalability in larger models.

\section{Limitations}
Despite our comprehensive analysis, several limitations remain.

\begin{itemize}
    \item \textbf{Computational Constraints:} Due to resource limitations, we could not extend our analysis to the neuron level, potentially missing finer-grained switching behaviors. Additionally, our role classification relies on a fixed activation threshold (\(\theta = 0.10\)) chosen based on the distribution of component contributions; while we examined that varying this threshold within a reasonable range does not alter the overall trends, its use may still introduce bias.

    \item \textbf{Model Checkpoints \& Variants:} We analyzed a subset of training snapshots, leaving gaps in tracking role transitions. Newer and larger versions of the model were released during this study, but incorporating them was infeasible. While related families such as Pythia~\cite{biderman:2023b} also offer multiple checkpoints, their underlying GPT-J architecture computes attention and FFN contributions in parallel, preventing the use of IFRs for circuit extraction. As a result, direct comparison with OLMo-7B via IFR is not possible, though such comparisons could yield additional insights in future work.

    \item \textbf{Dataset Scope \& Generalizability:} Our dataset focuses on factual recall in English, covering only location-based and name-based relations. Expanding to other domains, multilingual settings, and ambiguous queries would improve generalizability.

    \item \textbf{Interpretability Framework:} While IFRs efficiently trace knowledge circuits at billion-parameter scale, they may miss finer-grained interactions. Alternative techniques, such as activation patching or causal tracing, are computationally prohibitive to apply across 40 snapshots of a 7B-parameter model, so direct method-to-method comparisons were not feasible in this study.

    \item \textbf{Model Adaptability \& Downstream Implications:} While attention heads frequently transition roles, FFNs remain stable, but their long-term impact on fine-tuning, pruning, and continual learning is unclear. Investigating their adaptability could enhance optimization strategies.
\end{itemize}

Future work should address these limitations by incorporating more diverse datasets, additional model variants, and alternative interpretability techniques to deepen our understanding of knowledge formation in LLMs.

\section*{Acknowledgements}
This research was supported by DFG (grant SCHU 2246/14-1).
We would like to thank the members of the SchützeLab and MaiNLP for their valuable feedback and discussions, and we also acknowledge the anonymous reviewers for their helpful suggestions.

\bibliography{references}

\begin{thebibliography}{50}
\providecommand{\natexlab}[1]{#1}

\bibitem[{Biderman et~al.(2023{\natexlab{a}})Biderman, Prashanth, Sutawika, Schoelkopf, Anthony, Purohit, and Raff}]{biderman:2023}
Stella Biderman, USVSN~Sai Prashanth, Lintang Sutawika, Hailey Schoelkopf, Quentin Anthony, Shivanshu Purohit, and Edward Raff. 2023{\natexlab{a}}.
\newblock \href {http://papers.nips.cc/paper\_files/paper/2023/hash/59404fb89d6194641c69ae99ecdf8f6d-Abstract-Conference.html} {Emergent and predictable memorization in large language models}.
\newblock In \emph{Advances in Neural Information Processing Systems 36: Annual Conference on Neural Information Processing Systems 2023, NeurIPS 2023, New Orleans, LA, USA, December 10 - 16, 2023}.

\bibitem[{Biderman et~al.(2023{\natexlab{b}})Biderman, Schoelkopf, Anthony, Bradley, O'Brien, Hallahan, Khan, Purohit, Prashanth, Raff, Skowron, Sutawika, and van~der Wal}]{biderman:2023b}
Stella Biderman, Hailey Schoelkopf, Quentin~Gregory Anthony, Herbie Bradley, Kyle O'Brien, Eric Hallahan, Mohammad~Aflah Khan, Shivanshu Purohit, USVSN~Sai Prashanth, Edward Raff, Aviya Skowron, Lintang Sutawika, and Oskar van~der Wal. 2023{\natexlab{b}}.
\newblock \href {https://proceedings.mlr.press/v202/biderman23a.html} {Pythia: {A} suite for analyzing large language models across training and scaling}.
\newblock In \emph{International Conference on Machine Learning, {ICML} 2023, 23-29 July 2023, Honolulu, Hawaii, {USA}}, volume 202 of \emph{Proceedings of Machine Learning Research}, pages 2397--2430. {PMLR}.

\bibitem[{Bricken et~al.(2023)Bricken, Templeton, Batson, Chen, Jermyn, Conerly, Turner, Anil, Denison, Askell, Lasenby, Wu, Kravec, Schiefer, Maxwell, Joseph, Hatfield-Dodds, Tamkin, Nguyen, McLean, Burke, Hume, Carter, Henighan, and Olah}]{bricken:2023}
Trenton Bricken, Adly Templeton, Joshua Batson, Brian Chen, Adam Jermyn, Tom Conerly, Nick Turner, Cem Anil, Carson Denison, Amanda Askell, Robert Lasenby, Yifan Wu, Shauna Kravec, Nicholas Schiefer, Tim Maxwell, Nicholas Joseph, Zac Hatfield-Dodds, Alex Tamkin, Karina Nguyen, Brayden McLean, Josiah~E. Burke, Tristan Hume, Shan Carter, Tom Henighan, and Christopher Olah. 2023.
\newblock Towards monosemanticity: Decomposing language models with dictionary learning.
\newblock \emph{Transformer Circuits Thread}.

\bibitem[{Chang et~al.(2024)Chang, Park, Ye, Yang, Seo, Chang, and Seo}]{chang:2024}
Hoyeon Chang, Jinho Park, Seonghyeon Ye, Sohee Yang, Youngkyung Seo, Du{-}Seong Chang, and Minjoon Seo. 2024.
\newblock \href {http://papers.nips.cc/paper\_files/paper/2024/hash/6fdf57c71bc1f1ee29014b8dc52e723f-Abstract-Conference.html} {How do large language models acquire factual knowledge during pretraining?}
\newblock In \emph{Advances in Neural Information Processing Systems 38: Annual Conference on Neural Information Processing Systems 2024, NeurIPS 2024, Vancouver, BC, Canada, December 10 - 15, 2024}.

\bibitem[{Chang et~al.(2023)Chang, Tu, and Bergen}]{chang:2023}
Tyler~A. Chang, Zhuowen Tu, and Benjamin~K. Bergen. 2023.
\newblock \href {https://doi.org/10.48550/ARXIV.2308.15419} {Characterizing learning curves during language model pre-training: Learning, forgetting, and stability}.
\newblock \emph{CoRR}, abs/2308.15419.

\bibitem[{Chen et~al.(2024)Chen, Shwartz{-}Ziv, Cho, Leavitt, and Saphra}]{chen:2024}
Angelica Chen, Ravid Shwartz{-}Ziv, Kyunghyun Cho, Matthew~L. Leavitt, and Naomi Saphra. 2024.
\newblock \href {https://openreview.net/forum?id=MO5PiKHELW} {Sudden drops in the loss: Syntax acquisition, phase transitions, and simplicity bias in mlms}.
\newblock In \emph{The Twelfth International Conference on Learning Representations, {ICLR} 2024, Vienna, Austria, May 7-11, 2024}. OpenReview.net.

\bibitem[{Chiang et~al.(2020)Chiang, Huang, and Lee}]{chiang:2020}
David~Cheng{-}Han Chiang, Sung{-}Feng Huang, and Hung{-}yi Lee. 2020.
\newblock \href {https://doi.org/10.18653/V1/2020.EMNLP-MAIN.553} {Pretrained language model embryology: The birth of {ALBERT}}.
\newblock In \emph{Proceedings of the 2020 Conference on Empirical Methods in Natural Language Processing, {EMNLP} 2020, Online, November 16-20, 2020}, pages 6813--6828. Association for Computational Linguistics.

\bibitem[{Chuang et~al.(2024)Chuang, Xie, Luo, Kim, Glass, and He}]{chuang:2024}
Yung{-}Sung Chuang, Yujia Xie, Hongyin Luo, Yoon Kim, James~R. Glass, and Pengcheng He. 2024.
\newblock \href {https://openreview.net/forum?id=Th6NyL07na} {Dola: Decoding by contrasting layers improves factuality in large language models}.
\newblock In \emph{The Twelfth International Conference on Learning Representations, {ICLR} 2024, Vienna, Austria, May 7-11, 2024}. OpenReview.net.

\bibitem[{Chughtai et~al.(2024)Chughtai, Cooney, and Nanda}]{chughtai:2024}
Bilal Chughtai, Alan Cooney, and Neel Nanda. 2024.
\newblock \href {https://doi.org/10.48550/ARXIV.2402.07321} {Summing up the facts: Additive mechanisms behind factual recall in llms}.
\newblock \emph{CoRR}, abs/2402.07321.

\bibitem[{Elazar et~al.(2021)Elazar, Kassner, Ravfogel, Ravichander, Hovy, Sch{\"{u}}tze, and Goldberg}]{elazar:2021}
Yanai Elazar, Nora Kassner, Shauli Ravfogel, Abhilasha Ravichander, Eduard~H. Hovy, Hinrich Sch{\"{u}}tze, and Yoav Goldberg. 2021.
\newblock \href {https://doi.org/10.1162/TACL\_A\_00410} {Measuring and improving consistency in pretrained language models}.
\newblock \emph{Trans. Assoc. Comput. Linguistics}, 9:1012--1031.

\bibitem[{Elhage et~al.(2021)Elhage, Nanda, Olsson, Henighan, Joseph, Mann, Askell, Bai, Chen, Conerly et~al.}]{elhage:2021}
Nelson Elhage, Neel Nanda, Catherine Olsson, Tom Henighan, Nicholas Joseph, Ben Mann, Amanda Askell, Yuntao Bai, Anna Chen, Tom Conerly, et~al. 2021.
\newblock A mathematical framework for transformer circuits.
\newblock \emph{Transformer Circuits Thread}, 1(1):12.

\bibitem[{Elhelo and Geva(2024)}]{elhelo:2024}
Amit Elhelo and Mor Geva. 2024.
\newblock \href {https://doi.org/10.48550/ARXIV.2412.11965} {Inferring functionality of attention heads from their parameters}.
\newblock \emph{CoRR}, abs/2412.11965.

\bibitem[{Ferrando et~al.(2022)Ferrando, G{\'{a}}llego, and Costa{-}juss{\`{a}}}]{ferrando:2022}
Javier Ferrando, Gerard~I. G{\'{a}}llego, and Marta~R. Costa{-}juss{\`{a}}. 2022.
\newblock \href {https://doi.org/10.18653/V1/2022.EMNLP-MAIN.595} {Measuring the mixing of contextual information in the transformer}.
\newblock In \emph{Proceedings of the 2022 Conference on Empirical Methods in Natural Language Processing, {EMNLP} 2022, Abu Dhabi, United Arab Emirates, December 7-11, 2022}, pages 8698--8714. Association for Computational Linguistics.

\bibitem[{Ferrando and Voita(2024)}]{ferrando:2024}
Javier Ferrando and Elena Voita. 2024.
\newblock \href {https://aclanthology.org/2024.emnlp-main.965} {Information flow routes: Automatically interpreting language models at scale}.
\newblock In \emph{Proceedings of the 2024 Conference on Empirical Methods in Natural Language Processing, {EMNLP} 2024, Miami, FL, USA, November 12-16, 2024}, pages 17432--17445. Association for Computational Linguistics.

\bibitem[{Geva et~al.(2023)Geva, Bastings, Filippova, and Globerson}]{geva:2023}
Mor Geva, Jasmijn Bastings, Katja Filippova, and Amir Globerson. 2023.
\newblock \href {https://doi.org/10.18653/V1/2023.EMNLP-MAIN.751} {Dissecting recall of factual associations in auto-regressive language models}.
\newblock In \emph{Proceedings of the 2023 Conference on Empirical Methods in Natural Language Processing, {EMNLP} 2023, Singapore, December 6-10, 2023}, pages 12216--12235. Association for Computational Linguistics.

\bibitem[{Geva et~al.(2022)Geva, Caciularu, Wang, and Goldberg}]{geva:2022}
Mor Geva, Avi Caciularu, Kevin~Ro Wang, and Yoav Goldberg. 2022.
\newblock \href {https://doi.org/10.18653/V1/2022.EMNLP-MAIN.3} {Transformer feed-forward layers build predictions by promoting concepts in the vocabulary space}.
\newblock In \emph{Proceedings of the 2022 Conference on Empirical Methods in Natural Language Processing, {EMNLP} 2022, Abu Dhabi, United Arab Emirates, December 7-11, 2022}, pages 30--45. Association for Computational Linguistics.

\bibitem[{Geva et~al.(2021)Geva, Schuster, Berant, and Levy}]{geva:2021}
Mor Geva, Roei Schuster, Jonathan Berant, and Omer Levy. 2021.
\newblock \href {https://doi.org/10.18653/V1/2021.EMNLP-MAIN.446} {Transformer feed-forward layers are key-value memories}.
\newblock In \emph{Proceedings of the 2021 Conference on Empirical Methods in Natural Language Processing, {EMNLP} 2021, Virtual Event / Punta Cana, Dominican Republic, 7-11 November, 2021}, pages 5484--5495. Association for Computational Linguistics.

\bibitem[{Gould et~al.(2024)Gould, Ong, Ogden, and Conmy}]{gould:2024}
Rhys Gould, Euan Ong, George Ogden, and Arthur Conmy. 2024.
\newblock \href {https://openreview.net/forum?id=kvcbV8KQsi} {Successor heads: Recurring, interpretable attention heads in the wild}.
\newblock In \emph{The Twelfth International Conference on Learning Representations, {ICLR} 2024, Vienna, Austria, May 7-11, 2024}. OpenReview.net.

\bibitem[{Groeneveld et~al.(2024)Groeneveld, Beltagy, Walsh, Bhagia, Kinney, Tafjord, Jha, Ivison, Magnusson, Wang, Arora, Atkinson, Authur, Chandu, Cohan, Dumas, Elazar, Gu, Hessel, Khot, Merrill, Morrison, Muennighoff, Naik, Nam, Peters, Pyatkin, Ravichander, Schwenk, Shah, Smith, Strubell, Subramani, Wortsman, Dasigi, Lambert, Richardson, Zettlemoyer, Dodge, Lo, Soldaini, Smith, and Hajishirzi}]{groeneveld:2024}
Dirk Groeneveld, Iz~Beltagy, Evan~Pete Walsh, Akshita Bhagia, Rodney Kinney, Oyvind Tafjord, Ananya~Harsh Jha, Hamish Ivison, Ian Magnusson, Yizhong Wang, Shane Arora, David Atkinson, Russell Authur, Khyathi~Raghavi Chandu, Arman Cohan, Jennifer Dumas, Yanai Elazar, Yuling Gu, Jack Hessel, Tushar Khot, William Merrill, Jacob Morrison, Niklas Muennighoff, Aakanksha Naik, Crystal Nam, Matthew~E. Peters, Valentina Pyatkin, Abhilasha Ravichander, Dustin Schwenk, Saurabh Shah, Will Smith, Emma Strubell, Nishant Subramani, Mitchell Wortsman, Pradeep Dasigi, Nathan Lambert, Kyle Richardson, Luke Zettlemoyer, Jesse Dodge, Kyle Lo, Luca Soldaini, Noah~A. Smith, and Hannaneh Hajishirzi. 2024.
\newblock \href {https://doi.org/10.18653/V1/2024.ACL-LONG.841} {Olmo: Accelerating the science of language models}.
\newblock In \emph{Proceedings of the 62nd Annual Meeting of the Association for Computational Linguistics (Volume 1: Long Papers), {ACL} 2024, Bangkok, Thailand, August 11-16, 2024}, pages 15789--15809. Association for Computational Linguistics.

\bibitem[{Hanna et~al.(2023)Hanna, Liu, and Variengien}]{hanna:2023}
Michael Hanna, Ollie Liu, and Alexandre Variengien. 2023.
\newblock \href {http://papers.nips.cc/paper\_files/paper/2023/hash/efbba7719cc5172d175240f24be11280-Abstract-Conference.html} {How does {GPT-2} compute greater-than?: Interpreting mathematical abilities in a pre-trained language model}.
\newblock In \emph{Advances in Neural Information Processing Systems 36: Annual Conference on Neural Information Processing Systems 2023, NeurIPS 2023, New Orleans, LA, USA, December 10 - 16, 2023}.

\bibitem[{Hanna et~al.(2024)Hanna, Pezzelle, and Belinkov}]{hanna:2024}
Michael Hanna, Sandro Pezzelle, and Yonatan Belinkov. 2024.
\newblock \href {https://doi.org/10.48550/ARXIV.2403.17806} {Have faith in faithfulness: Going beyond circuit overlap when finding model mechanisms}.
\newblock \emph{CoRR}, abs/2403.17806.

\bibitem[{Haviv et~al.(2023)Haviv, Cohen, Gidron, Schuster, Goldberg, and Geva}]{haviv:2023}
Adi Haviv, Ido Cohen, Jacob Gidron, Roei Schuster, Yoav Goldberg, and Mor Geva. 2023.
\newblock \href {https://doi.org/10.18653/V1/2023.EACL-MAIN.19} {Understanding transformer memorization recall through idioms}.
\newblock In \emph{Proceedings of the 17th Conference of the European Chapter of the Association for Computational Linguistics, {EACL} 2023, Dubrovnik, Croatia, May 2-6, 2023}, pages 248--264. Association for Computational Linguistics.

\bibitem[{Hernandez et~al.(2024)Hernandez, Sharma, Haklay, Meng, Wattenberg, Andreas, Belinkov, and Bau}]{hernandez:2024}
Evan Hernandez, Arnab~Sen Sharma, Tal Haklay, Kevin Meng, Martin Wattenberg, Jacob Andreas, Yonatan Belinkov, and David Bau. 2024.
\newblock \href {https://openreview.net/forum?id=w7LU2s14kE} {Linearity of relation decoding in transformer language models}.
\newblock In \emph{The Twelfth International Conference on Learning Representations, {ICLR} 2024, Vienna, Austria, May 7-11, 2024}. OpenReview.net.

\bibitem[{Hu et~al.(2023)Hu, Chen, Saphra, and Cho}]{hu:2023}
Michael~Y. Hu, Angelica Chen, Naomi Saphra, and Kyunghyun Cho. 2023.
\newblock \href {https://openreview.net/forum?id=NE2xXWo0LF} {Latent state models of training dynamics}.
\newblock \emph{Trans. Mach. Learn. Res.}, 2023.

\bibitem[{Lieberum et~al.(2023)Lieberum, Rahtz, Kram{\'{a}}r, Nanda, Irving, Shah, and Mikulik}]{lieberum:2023}
Tom Lieberum, Matthew Rahtz, J{\'{a}}nos Kram{\'{a}}r, Neel Nanda, Geoffrey Irving, Rohin Shah, and Vladimir Mikulik. 2023.
\newblock \href {https://doi.org/10.48550/ARXIV.2307.09458} {Does circuit analysis interpretability scale? evidence from multiple choice capabilities in chinchilla}.
\newblock \emph{CoRR}, abs/2307.09458.

\bibitem[{Liu et~al.(2021)Liu, Wang, Kasai, Hajishirzi, and Smith}]{liu:2021}
Zeyu Liu, Yizhong Wang, Jungo Kasai, Hannaneh Hajishirzi, and Noah~A. Smith. 2021.
\newblock \href {https://doi.org/10.18653/V1/2021.FINDINGS-EMNLP.71} {Probing across time: What does roberta know and when?}
\newblock In \emph{Findings of the Association for Computational Linguistics: {EMNLP} 2021, Virtual Event / Punta Cana, Dominican Republic, 16-20 November, 2021}, pages 820--842. Association for Computational Linguistics.

\bibitem[{McDougall et~al.(2023)McDougall, Conmy, Rushing, McGrath, and Nanda}]{mcdougall:2023}
Callum McDougall, Arthur Conmy, Cody Rushing, Thomas McGrath, and Neel Nanda. 2023.
\newblock \href {https://doi.org/10.48550/ARXIV.2310.04625} {Copy suppression: Comprehensively understanding an attention head}.
\newblock \emph{CoRR}, abs/2310.04625.

\bibitem[{McGrath et~al.(2023)McGrath, Rahtz, Kram{\'{a}}r, Mikulik, and Legg}]{mcgrath:2023}
Thomas McGrath, Matthew Rahtz, J{\'{a}}nos Kram{\'{a}}r, Vladimir Mikulik, and Shane Legg. 2023.
\newblock \href {https://doi.org/10.48550/ARXIV.2307.15771} {The hydra effect: Emergent self-repair in language model computations}.
\newblock \emph{CoRR}, abs/2307.15771.

\bibitem[{Meng et~al.(2022)Meng, Bau, Andonian, and Belinkov}]{meng:2022}
Kevin Meng, David Bau, Alex Andonian, and Yonatan Belinkov. 2022.
\newblock \href {http://papers.nips.cc/paper\_files/paper/2022/hash/6f1d43d5a82a37e89b0665b33bf3a182-Abstract-Conference.html} {Locating and editing factual associations in {GPT}}.
\newblock In \emph{Advances in Neural Information Processing Systems 35: Annual Conference on Neural Information Processing Systems 2022, NeurIPS 2022, New Orleans, LA, USA, November 28 - December 9, 2022}.

\bibitem[{Merullo et~al.(2024)Merullo, Eickhoff, and Pavlick}]{merullo:2024}
Jack Merullo, Carsten Eickhoff, and Ellie Pavlick. 2024.
\newblock \href {https://openreview.net/forum?id=fpoAYV6Wsk} {Circuit component reuse across tasks in transformer language models}.
\newblock In \emph{The Twelfth International Conference on Learning Representations, {ICLR} 2024, Vienna, Austria, May 7-11, 2024}. OpenReview.net.

\bibitem[{Mondorf et~al.(2024)Mondorf, Wold, and Plank}]{mondorf:2024}
Philipp Mondorf, Sondre Wold, and Barbara Plank. 2024.
\newblock \href {https://doi.org/10.48550/ARXIV.2410.01434} {Circuit compositions: Exploring modular structures in transformer-based language models}.
\newblock \emph{CoRR}, abs/2410.01434.

\bibitem[{M{\"{u}}ller{-}Eberstein et~al.(2023)M{\"{u}}ller{-}Eberstein, van~der Goot, Plank, and Titov}]{muller-eberstein:2023}
Max M{\"{u}}ller{-}Eberstein, Rob van~der Goot, Barbara Plank, and Ivan Titov. 2023.
\newblock \href {https://doi.org/10.18653/V1/2023.FINDINGS-EMNLP.879} {Subspace chronicles: How linguistic information emerges, shifts and interacts during language model training}.
\newblock In \emph{Findings of the Association for Computational Linguistics: {EMNLP} 2023, Singapore, December 6-10, 2023}, pages 13190--13208. Association for Computational Linguistics.

\bibitem[{Nanda et~al.(2023)Nanda, Chan, Lieberum, Smith, and Steinhardt}]{nanda:2023}
Neel Nanda, Lawrence Chan, Tom Lieberum, Jess Smith, and Jacob Steinhardt. 2023.
\newblock \href {https://openreview.net/forum?id=9XFSbDPmdW} {Progress measures for grokking via mechanistic interpretability}.
\newblock In \emph{The Eleventh International Conference on Learning Representations, {ICLR} 2023, Kigali, Rwanda, May 1-5, 2023}. OpenReview.net.

\bibitem[{Neo et~al.(2024)Neo, Cohen, and Barez}]{neo:2024}
Clement Neo, Shay~B. Cohen, and Fazl Barez. 2024.
\newblock \href {https://aclanthology.org/2024.emnlp-main.930} {Interpreting context look-ups in transformers: Investigating attention-mlp interactions}.
\newblock In \emph{Proceedings of the 2024 Conference on Empirical Methods in Natural Language Processing, {EMNLP} 2024, Miami, FL, USA, November 12-16, 2024}, pages 16681--16697. Association for Computational Linguistics.

\bibitem[{Olah et~al.(2020)Olah, Cammarata, Schubert, Goh, Petrov, and Carter}]{olah:2020}
Chris Olah, Nick Cammarata, Ludwig Schubert, Gabriel Goh, Michael Petrov, and Shan Carter. 2020.
\newblock Zoom in: An introduction to circuits.
\newblock \emph{Distill}, 5(3):e00024--001.

\bibitem[{Olsson et~al.(2022)Olsson, Elhage, Nanda, Joseph, DasSarma, Henighan, Mann, Askell, Bai, Chen, Conerly, Drain, Ganguli, Hatfield{-}Dodds, Hernandez, Johnston, Jones, Kernion, Lovitt, Ndousse, Amodei, Brown, Clark, Kaplan, McCandlish, and Olah}]{olsson:2022}
Catherine Olsson, Nelson Elhage, Neel Nanda, Nicholas Joseph, Nova DasSarma, Tom Henighan, Ben Mann, Amanda Askell, Yuntao Bai, Anna Chen, Tom Conerly, Dawn Drain, Deep Ganguli, Zac Hatfield{-}Dodds, Danny Hernandez, Scott Johnston, Andy Jones, Jackson Kernion, Liane Lovitt, Kamal Ndousse, Dario Amodei, Tom Brown, Jack Clark, Jared Kaplan, Sam McCandlish, and Chris Olah. 2022.
\newblock \href {https://doi.org/10.48550/ARXIV.2209.11895} {In-context learning and induction heads}.
\newblock \emph{CoRR}, abs/2209.11895.

\bibitem[{Ortu et~al.(2024)Ortu, Jin, Doimo, Sachan, Cazzaniga, and Sch{\"{o}}lkopf}]{ortu:2024}
Francesco Ortu, Zhijing Jin, Diego Doimo, Mrinmaya Sachan, Alberto Cazzaniga, and Bernhard Sch{\"{o}}lkopf. 2024.
\newblock \href {https://doi.org/10.18653/V1/2024.ACL-LONG.458} {Competition of mechanisms: Tracing how language models handle facts and counterfactuals}.
\newblock In \emph{Proceedings of the 62nd Annual Meeting of the Association for Computational Linguistics (Volume 1: Long Papers), {ACL} 2024, Bangkok, Thailand, August 11-16, 2024}, pages 8420--8436. Association for Computational Linguistics.

\bibitem[{Rogers et~al.(2020)Rogers, Kovaleva, and Rumshisky}]{rogers:2020}
Anna Rogers, Olga Kovaleva, and Anna Rumshisky. 2020.
\newblock \href {https://doi.org/10.1162/TACL\_A\_00349} {A primer in bertology: What we know about how {BERT} works}.
\newblock \emph{Trans. Assoc. Comput. Linguistics}, 8:842--866.

\bibitem[{Rushing and Nanda(2024)}]{rushing:2024}
Cody Rushing and Neel Nanda. 2024.
\newblock \href {https://openreview.net/forum?id=5ZwEifshyo} {Explorations of self-repair in language models}.
\newblock In \emph{Forty-first International Conference on Machine Learning, {ICML} 2024, Vienna, Austria, July 21-27, 2024}. OpenReview.net.

\bibitem[{Singh et~al.(2024)Singh, Moskovitz, Hill, Chan, and Saxe}]{singh:2024}
Aaditya~K. Singh, Ted Moskovitz, Felix Hill, Stephanie C.~Y. Chan, and Andrew~M. Saxe. 2024.
\newblock \href {https://openreview.net/forum?id=O8rrXl71D5} {What needs to go right for an induction head? {A} mechanistic study of in-context learning circuits and their formation}.
\newblock In \emph{Forty-first International Conference on Machine Learning, {ICML} 2024, Vienna, Austria, July 21-27, 2024}. OpenReview.net.

\bibitem[{Stoehr et~al.(2024)Stoehr, Gordon, Zhang, and Lewis}]{stoehr:2024}
Niklas Stoehr, Mitchell Gordon, Chiyuan Zhang, and Owen Lewis. 2024.
\newblock \href {https://doi.org/10.48550/ARXIV.2403.19851} {Localizing paragraph memorization in language models}.
\newblock \emph{CoRR}, abs/2403.19851.

\bibitem[{Sundararajan et~al.(2017)Sundararajan, Taly, and Yan}]{sundararajan:2017}
Mukund Sundararajan, Ankur Taly, and Qiqi Yan. 2017.
\newblock \href {http://proceedings.mlr.press/v70/sundararajan17a.html} {Axiomatic attribution for deep networks}.
\newblock In \emph{Proceedings of the 34th International Conference on Machine Learning, {ICML} 2017, Sydney, NSW, Australia, 6-11 August 2017}, volume~70 of \emph{Proceedings of Machine Learning Research}, pages 3319--3328. {PMLR}.

\bibitem[{Tian et~al.(2024)Tian, Wang, Zhang, Chen, and Du}]{tian:2024}
Yuandong Tian, Yiping Wang, Zhenyu Zhang, Beidi Chen, and Simon~Shaolei Du. 2024.
\newblock \href {https://openreview.net/forum?id=LbJqRGNYCf} {Joma: Demystifying multilayer transformers via joint dynamics of {MLP} and attention}.
\newblock In \emph{The Twelfth International Conference on Learning Representations, {ICLR} 2024, Vienna, Austria, May 7-11, 2024}. OpenReview.net.

\bibitem[{Tigges et~al.(2024)Tigges, Hanna, Yu, and Biderman}]{tigges:2024}
Curt Tigges, Michael Hanna, Qinan Yu, and Stella Biderman. 2024.
\newblock \href {http://papers.nips.cc/paper\_files/paper/2024/hash/47c7edadfee365b394b2a3bd416048da-Abstract-Conference.html} {{LLM} circuit analyses are consistent across training and scale}.
\newblock In \emph{Advances in Neural Information Processing Systems 38: Annual Conference on Neural Information Processing Systems 2024, NeurIPS 2024, Vancouver, BC, Canada, December 10 - 15, 2024}.

\bibitem[{Tigges et~al.(2023)Tigges, Hollinsworth, Geiger, and Nanda}]{tigges:2023}
Curt Tigges, Oskar~John Hollinsworth, Atticus Geiger, and Neel Nanda. 2023.
\newblock \href {https://doi.org/10.48550/ARXIV.2310.15154} {Linear representations of sentiment in large language models}.
\newblock \emph{CoRR}, abs/2310.15154.

\bibitem[{Variengien and Winsor(2023)}]{variengien:2023}
Alexandre Variengien and Eric Winsor. 2023.
\newblock \href {https://doi.org/10.48550/ARXIV.2312.10091} {Look before you leap: {A} universal emergent decomposition of retrieval tasks in language models}.
\newblock \emph{CoRR}, abs/2312.10091.

\bibitem[{Varma et~al.(2023)Varma, Shah, Kenton, Kram{\'{a}}r, and Kumar}]{varma:2023}
Vikrant Varma, Rohin Shah, Zachary Kenton, J{\'{a}}nos Kram{\'{a}}r, and Ramana Kumar. 2023.
\newblock \href {https://doi.org/10.48550/ARXIV.2309.02390} {Explaining grokking through circuit efficiency}.
\newblock \emph{CoRR}, abs/2309.02390.

\bibitem[{Wang et~al.(2023)Wang, Variengien, Conmy, Shlegeris, and Steinhardt}]{wang:2023}
Kevin~Ro Wang, Alexandre Variengien, Arthur Conmy, Buck Shlegeris, and Jacob Steinhardt. 2023.
\newblock \href {https://openreview.net/forum?id=NpsVSN6o4ul} {Interpretability in the wild: a circuit for indirect object identification in {GPT-2} small}.
\newblock In \emph{The Eleventh International Conference on Learning Representations, {ICLR} 2023, Kigali, Rwanda, May 1-5, 2023}. OpenReview.net.

\bibitem[{Xia et~al.(2023)Xia, Artetxe, Zhou, Lin, Pasunuru, Chen, Zettlemoyer, and Stoyanov}]{xia:2023}
Mengzhou Xia, Mikel Artetxe, Chunting Zhou, Xi~Victoria Lin, Ramakanth Pasunuru, Danqi Chen, Luke Zettlemoyer, and Veselin Stoyanov. 2023.
\newblock \href {https://doi.org/10.18653/V1/2023.ACL-LONG.767} {Training trajectories of language models across scales}.
\newblock In \emph{Proceedings of the 61st Annual Meeting of the Association for Computational Linguistics (Volume 1: Long Papers), {ACL} 2023, Toronto, Canada, July 9-14, 2023}, pages 13711--13738. Association for Computational Linguistics.

\bibitem[{Yu and Ananiadou(2024)}]{yu:2024}
Zeping Yu and Sophia Ananiadou. 2024.
\newblock \href {https://aclanthology.org/2024.emnlp-main.191} {Neuron-level knowledge attribution in large language models}.
\newblock In \emph{Proceedings of the 2024 Conference on Empirical Methods in Natural Language Processing, {EMNLP} 2024, Miami, FL, USA, November 12-16, 2024}, pages 3267--3280. Association for Computational Linguistics.

\end{thebibliography}

\clearpage 

\appendix
\pagebreak

\section{Implementation Details}
All datasets are in English. We employed AI assistants to improve the visual appeal and readability of both our data visualizations and certain sections of the text. Our setup involved an NVIDIA RTX A6000 alongside eight NVIDIA HGX A100-80x4-mig GPUs, which were used for inferring OLMo and extracting the circuits detailed in this work. Reproducing our full analysis and experiments takes about 24 hours for OLMo-7B using these eight GPUs.

\section{Dataset Construction}
\label{app:dataset-construction-pipeline}
\begin{enumerate} 
    \item \textbf{Prompt Template Design and Fact Collection}: For each of the 10 relations, we compiled 10 prompt templates. These prompts were paired with factual examples to serve as inputs for model evaluation.
    
    \item \textbf{Template Evaluation and Selection}: We tested all prompt templates with various factual inputs and determined the best-performing one for each relation. The evaluation was based on: \begin{itemize} 
        \item The \textbf{average probability} of the facts where the \textbf{first token} is correct. 
        \item The \textbf{reliability score} of the \textbf{second token}, which is calculated as the ratio of valid tokens for the second token (tokens with a probability less than 10\%) divided by the total amount of facts.
    \end{itemize} 
    Prompts were ranked based on a combined score derived from the average probability of the first token and the reliability of the second token. This scoring ensured that the prompts produced semantically accurate outputs, not merely syntactic completions.
    
    \item \textbf{Fact Reliability Validation}: Using the best-performing template for each relation, reliable facts were identified by ensuring that: 
    \begin{itemize} 
        \item The \textbf{top-1 token} is correct with a probability above 75\%. 
        \item The \textbf{second token} has a probability below 10\%. 
    \end{itemize} 
    This approach reduced reliance on syntactic biases and confirmed the semantic validity of the model's predictions. 
    \item \textbf{Final Dataset Generation}: For each relation, the dataset was finalized by pairing the best-performing prompt template with the set of validated, reliable facts.
\end{enumerate}

The resulting dataset includes 160 facts over 10 relations, each with a single best-performing prompt template and a curated collection of reliable facts validated for high accuracy and consistency. Prompts and Facts were validated using the main model to establish a reliable baseline for tracking knowledge evolution.

To ensure the robustness of our dataset, we prioritized semantically meaningful continuations over syntactic ones by evaluating both the first and second token probabilities. A strict scoring framework ensured that the top-1 token accurately reflected the correct answer while minimizing interference from alternative tokens. By combining insights from prior datasets with meticulous manual curation, we created a high-quality resource for probing factual knowledge. 

\newpage

\section{Accuracy Plots per Relation}
\label{app:accuracy-plots-per-relation}
\begin{figure}[h]
    \centering
    \includegraphics[width=\textwidth]{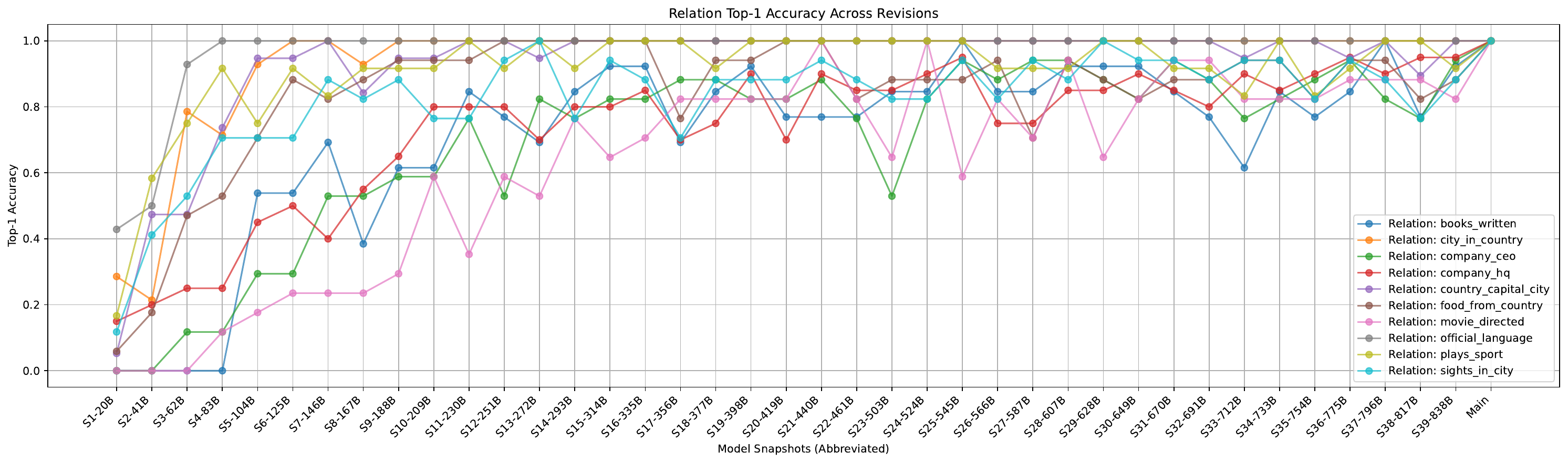}
    \begin{minipage}{\textwidth} 
        \centering
        \caption{Top-1 accuracy across different revisions of the Olmo model. Snapshots (\(S_X\)-\(YB\)) represent training checkpoints taken at 5000-step intervals, where \(Y\) indicates the number of tokens processed in billions.}
        \label{fig:top-1-acc-relation}
    \end{minipage}
\end{figure}

\begin{figure}[h]
    \centering
    \includegraphics[width=\textwidth]{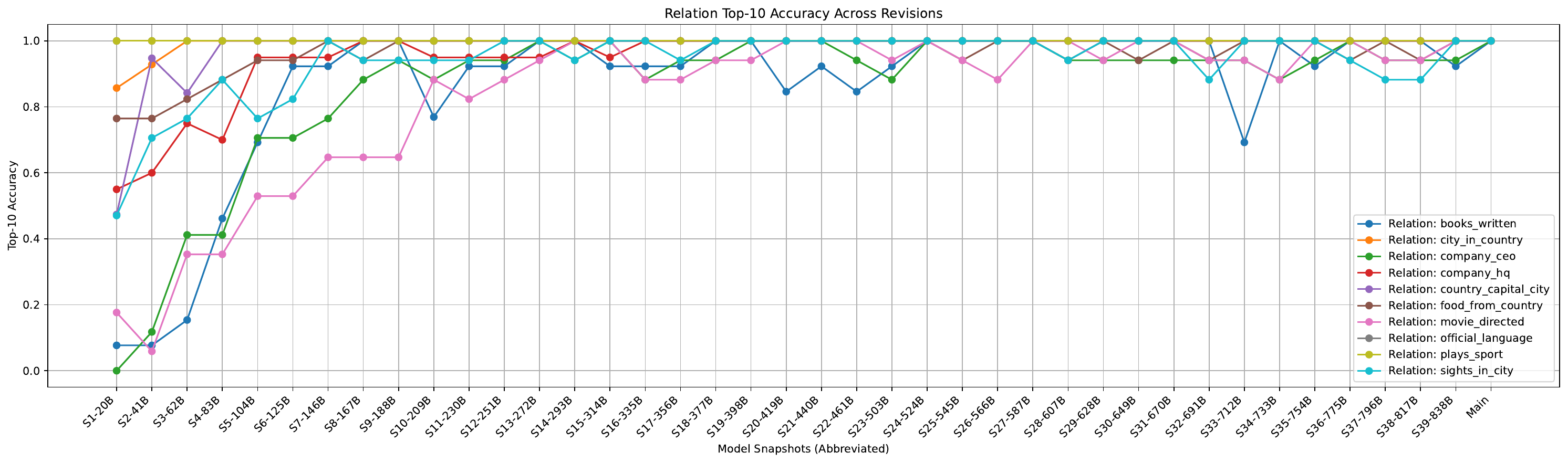}
    \begin{minipage}{\textwidth}  
        \centering
        \caption{Top-10 accuracy across different revisions of the Olmo model. Snapshots (\(S_X\)-\(YB\)) represent training checkpoints taken at 5000-step intervals, where \(Y\) indicates the number of tokens processed in billions.}
        \label{fig:top-10-acc-relation}
    \end{minipage}
\end{figure}

To complement the aggregated accuracy results in Section~\ref{sec:accuracy}, we present relation-level accuracy trends for top-1 and top-10 metrics across different model snapshots. These plots illustrate that NAME-based relations require significantly more training to achieve high accuracy compared to LOC relations (Figures~\ref{fig:top-1-acc-relation} and \ref{fig:top-10-acc-relation}). Among NAME relations, MOVIE\_DIRECTED is the most challenging, requiring approximately S10 to reach high top-10 accuracy, while COMPANY\_CEO and BOOKS\_WRITTEN also exhibit slower convergence. In contrast, LOC relations such as PLAYS\_SPORT and CITY\_IN\_COUNTRY are learned much faster.

For top-1 accuracy, OFFICIAL\_LANGUAGE is the first relation to reach 100\%, achieving this milestone at S4, whereas in the top-10 metric, it already attains 100\% as early as S1. This suggests that while correct answers are recognized among the top candidates from the beginning, ranking them correctly requires additional training.

\clearpage

\section{Counts of Relations}
\label{app:counts}
We obtained these statistics by querying the Infini-Gram API\footnote{\href{https://infini-gram.io/}{https://infini-gram.io/}} for every subject and target entity in our probing dataset. For each relation, we summed the API’s reported corpus frequencies of its subjects and targets to produce a “total count”, and then divided by the number of unique entities to compute an average per entity. We also recorded the number of unique entities per relation to capture its breadth. Relations were partitioned into two groups: NAME (name-based relations) and LOC (all location-based and miscellaneous relations) and the group-level table aggregates their totals, per-relation averages, and per-entity averages.

\begin{table}[h]
    \noindent
    % Create a "column‐width" box that holds both the resized tabular and its caption
    \begin{minipage}{\columnwidth}
      \centering
      \footnotesize
      \renewcommand{\arraystretch}{0.8}
      \setlength{\tabcolsep}{4pt}
  
      % Resize the tabular to exactly the width of this minipage
      \resizebox{\columnwidth}{!}{%
        \begin{tabular}{@{}lrrr@{}}
          \toprule
          \textbf{Relation}
            & \textbf{Total Count}
              & \textbf{\# Entities}
                & \textbf{Avg Count/Entity} \\
          \midrule
          \multicolumn{4}{@{}l}{\bfseries\scshape LOC relations} \\
          city\_in\_country      & 2\,636\,320  & 27  &  97\,641.48 \\
          company\_hq            & 1\,479\,816  & 34  &  43\,524.00 \\
          country\_capital\_city & 4\,884\,047  & 38  & 128\,527.55 \\
          food\_from\_country    & 2\,198\,553  & 30  &  73\,285.10 \\
          official\_language     & 4\,931\,312  & 26  & 189\,665.85 \\
          plays\_sport           &   505\,354   & 16  &  31\,584.62 \\
          sights\_in\_city       & 2\,163\,938  & 33  &  65\,573.88 \\
          \midrule
          \multicolumn{4}{@{}l}{\bfseries\scshape NAME relations} \\
          books\_written         &    12\,705   & 24  &     529.38   \\
          company\_ceo           &   146\,780   & 34  &   4\,317.06  \\
          movie\_directed        &    20\,567   & 32  &     642.72   \\
          \midrule
          LOC Summary            & 18\,799\,340 & 204 &  92\,153.63  \\
          NAME Summary           &   180\,052   &  90 &   2\,000.58  \\
          \midrule
          Overall Total          & 18\,979\,392 & 294 &  64\,555.75  \\
          \bottomrule
        \end{tabular}%
      } % end resizebox
  
      % Now the caption will span exactly \columnwidth
      \captionsetup{width=\columnwidth}
      \caption{%
        Per‐relation and group‐summary Infini‐Gram frequency statistics.
        We list each relation’s total corpus frequency (sum of subject + target counts),
        the number of unique entities (subjects + targets), and the average frequency per entity;
        then report LOC‐ and NAME‐group aggregates and the overall total.%
      }
      \label{tab:combined-stats}
  
    \end{minipage}
  \end{table}

\clearpage

\section{Relation-Level Component Counts and IoU Values}
\label{app:relation-level-counts-iou}
\subsection{Attention Heads}
\begin{figure}[h!]
    \centering
    \includegraphics[width=\textwidth]{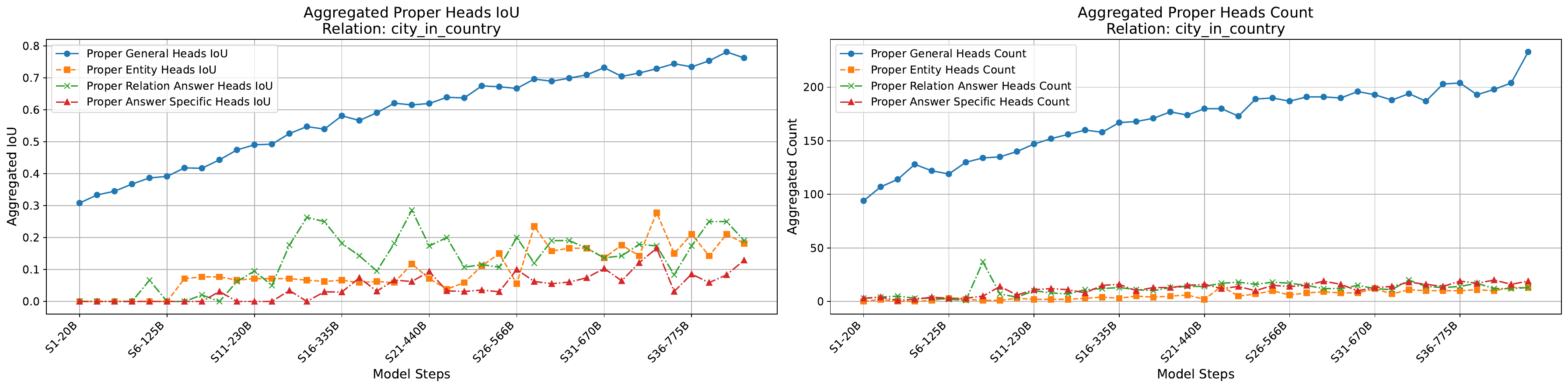}
    
    \vspace{0.3cm}
    \includegraphics[width=\textwidth]{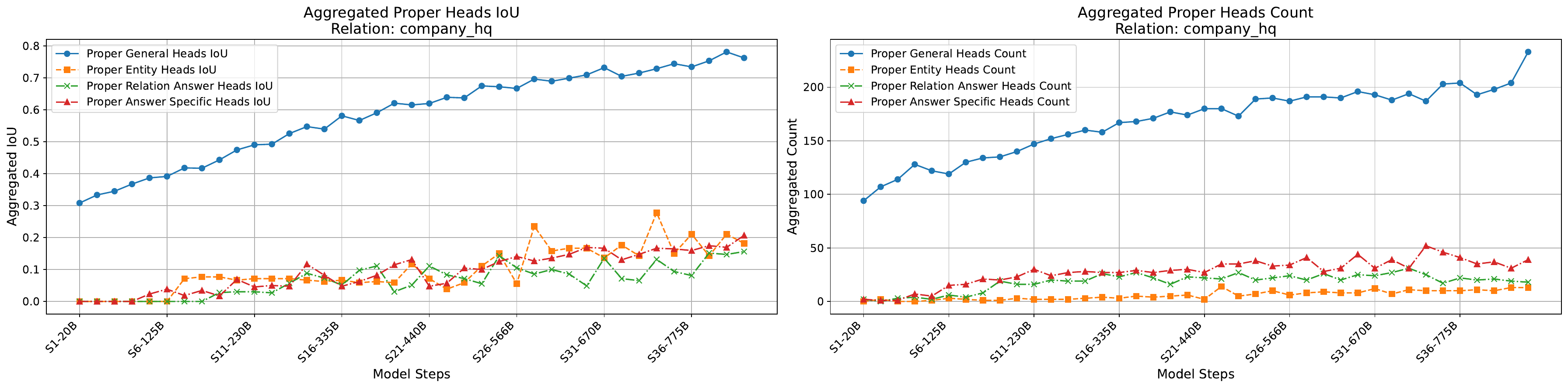}
    
    \vspace{0.3cm}
    \includegraphics[width=\textwidth]{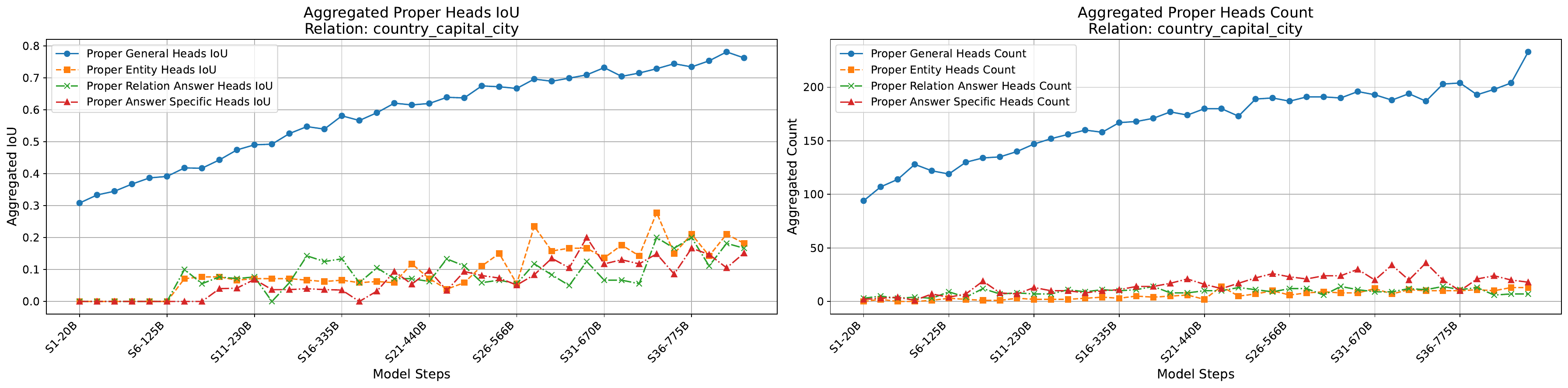}
    
    \vspace{0.3cm}
    \includegraphics[width=\textwidth]{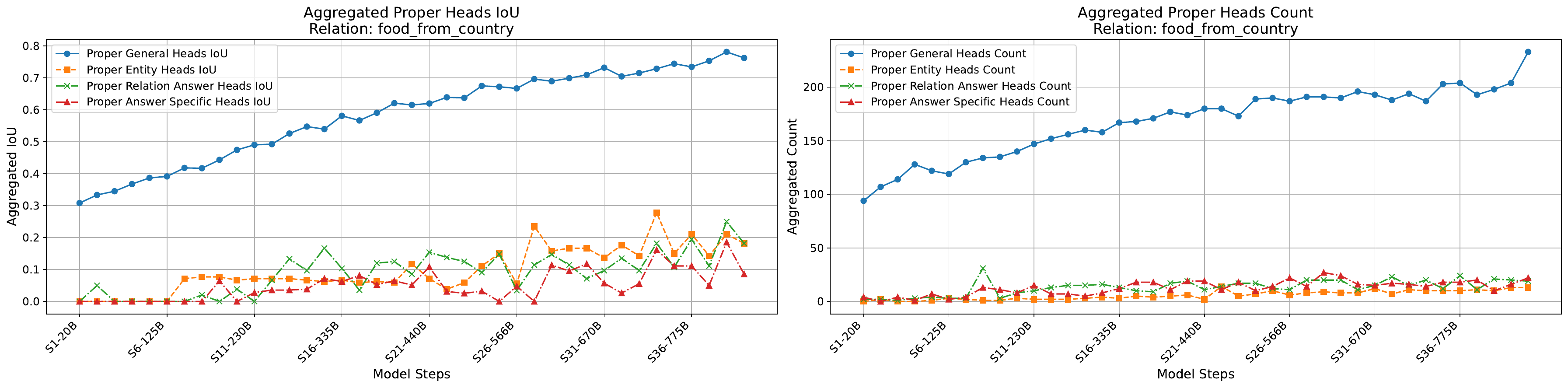}
    
    \vspace{0.3cm}
    \includegraphics[width=\textwidth]{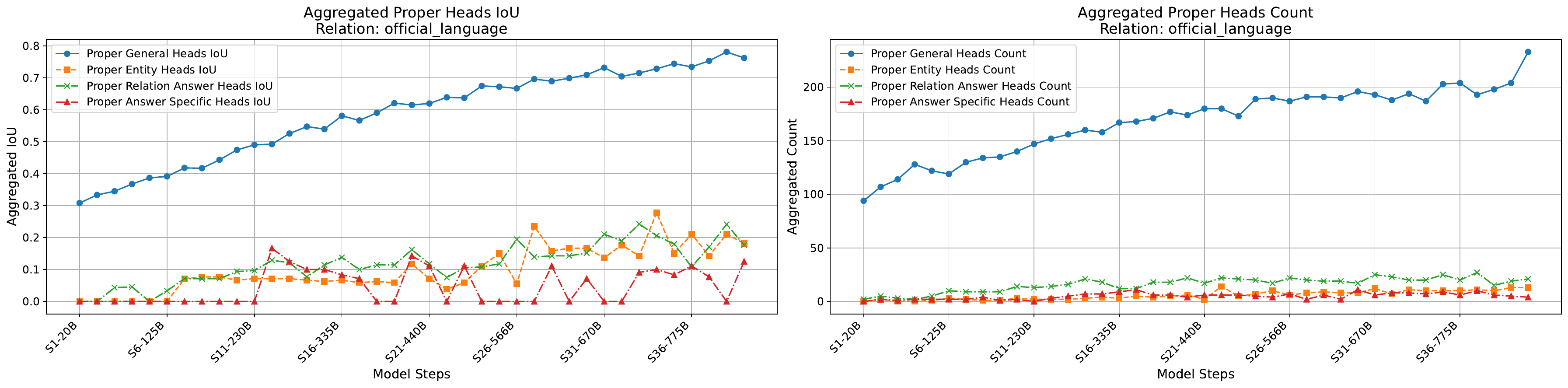}
    
    \begin{minipage}{\textwidth}  % Ensures full width
        \centering
        \caption{Relation-level head counts and IoU values.}
        \label{fig:all-subplots}
    \end{minipage}

\end{figure}

\clearpage
\begin{figure}[h!]
    \centering
    \ContinuedFloat % Continues numbering from the previous figure
    \includegraphics[width=\textwidth]{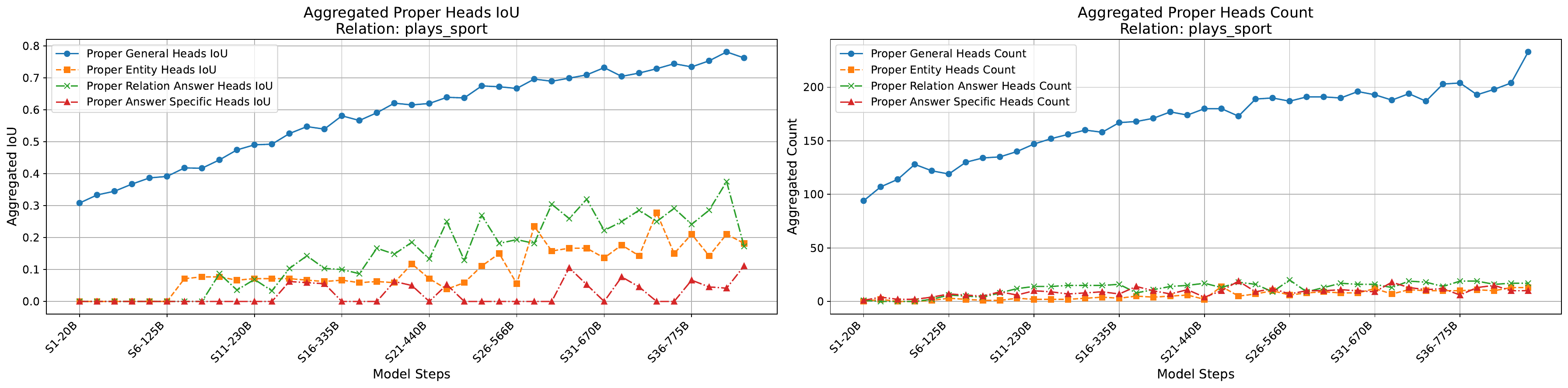}
    
    \vspace{0.3cm}
    \includegraphics[width=\textwidth]{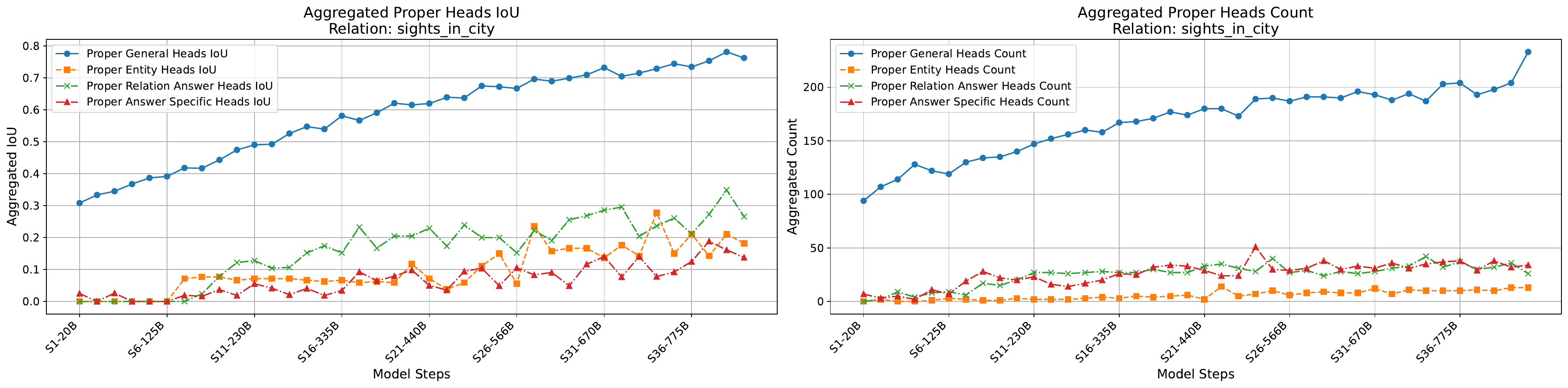}
    
    \vspace{0.3cm}
    \includegraphics[width=\textwidth]{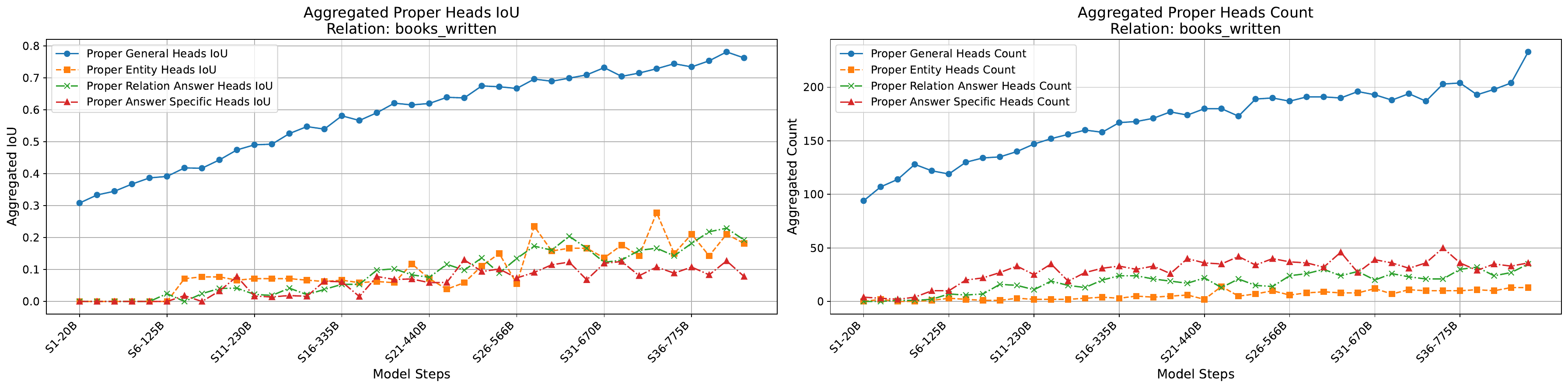}
    
    \vspace{0.3cm}
    \includegraphics[width=\textwidth]{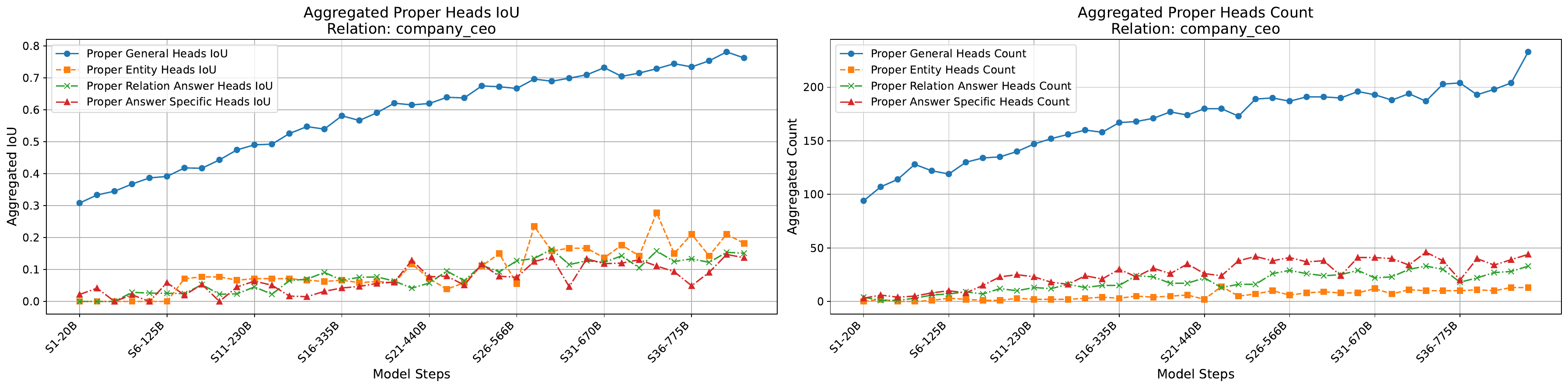}
    
    \vspace{0.3cm}
    \includegraphics[width=\textwidth]{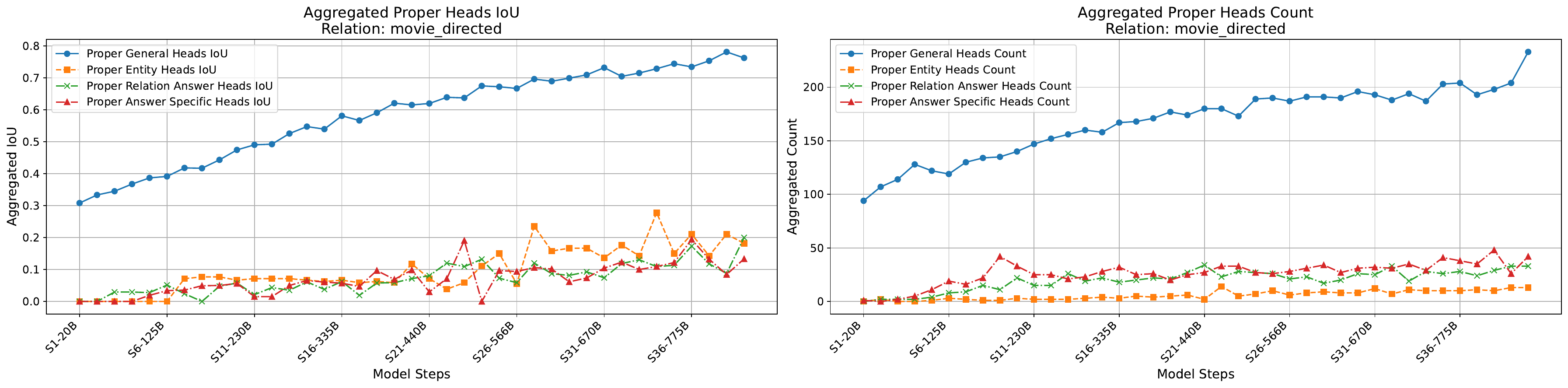}
    \begin{minipage}{\textwidth}  % Ensures full width
        \centering
        \caption{(continued) Relation-level head counts and IoU values.}
    \end{minipage}
\end{figure}

\clearpage

\subsection{Feed Forward Networks}

\begin{figure}[h!]
    \centering
    \includegraphics[width=\textwidth]{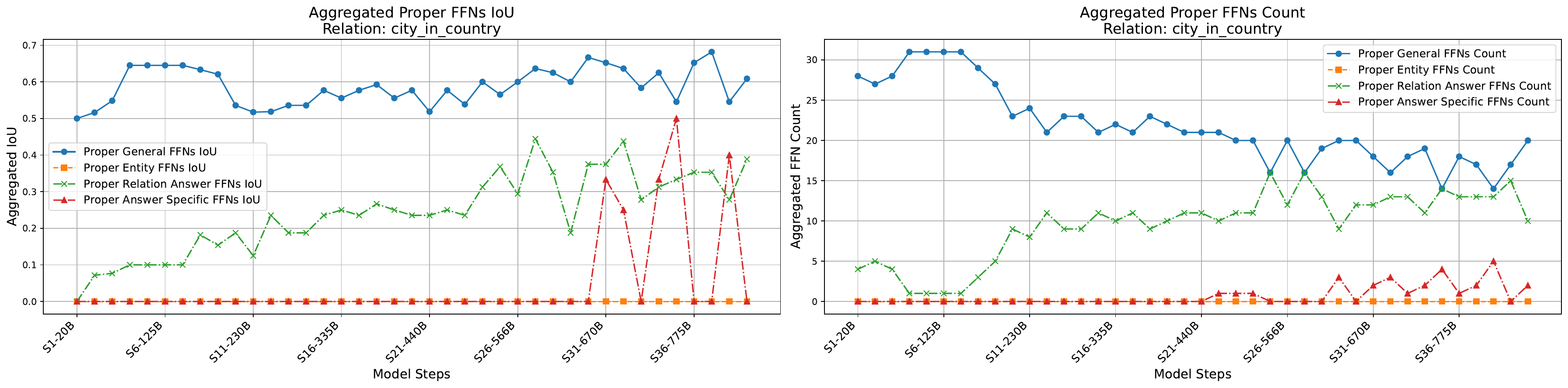}
    
    \vspace{0.3cm}
    \includegraphics[width=\textwidth]{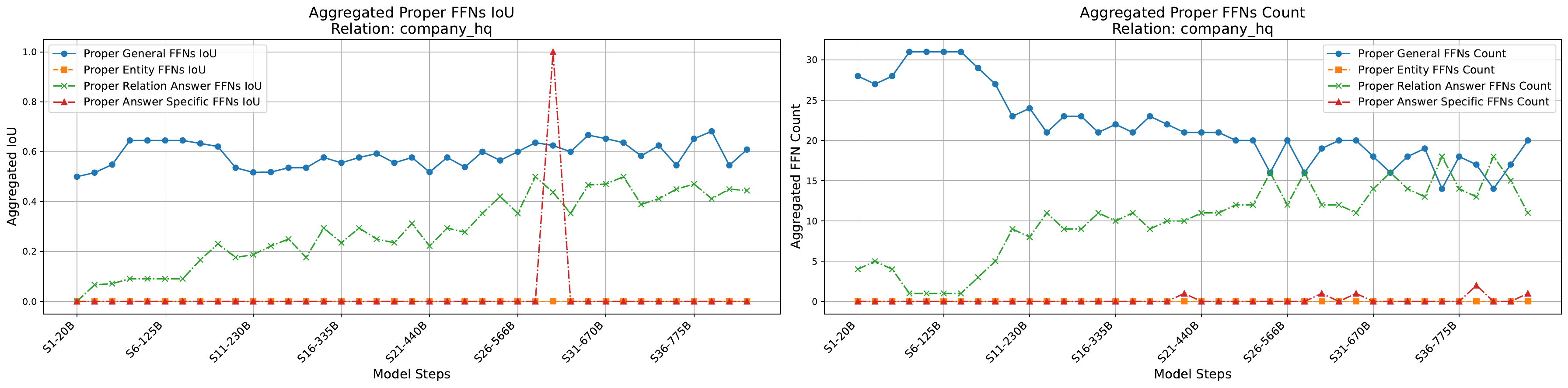}
    
    \vspace{0.3cm}
    \includegraphics[width=\textwidth]{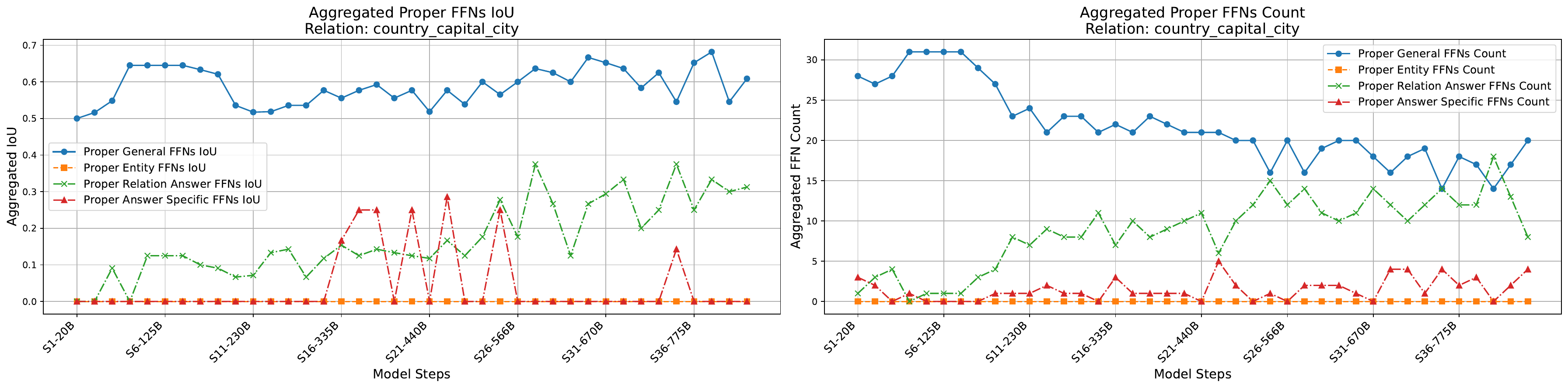}
    
    \vspace{0.3cm}
    \includegraphics[width=\textwidth]{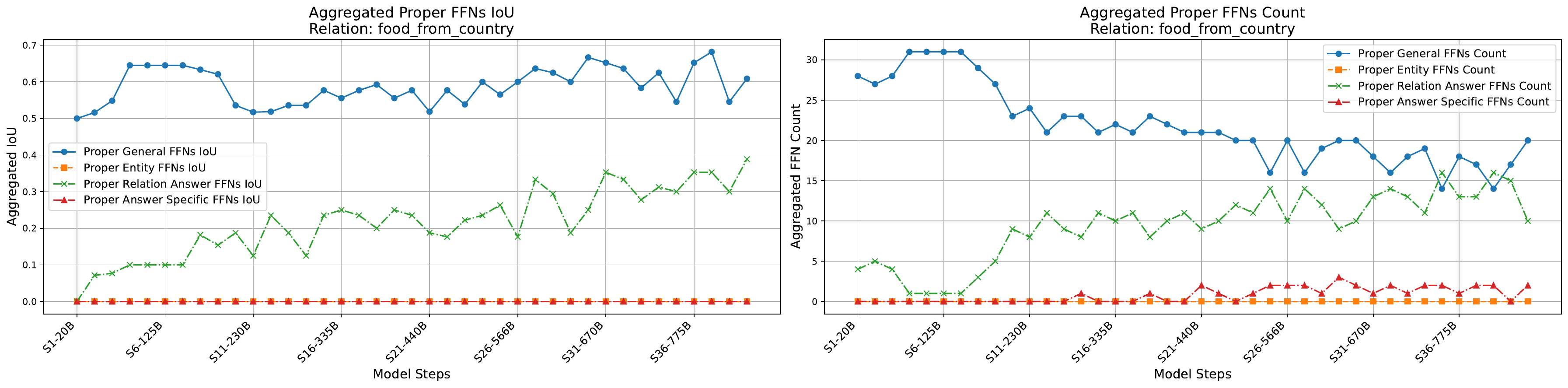}
    
    \vspace{0.3cm}
    \includegraphics[width=\textwidth]{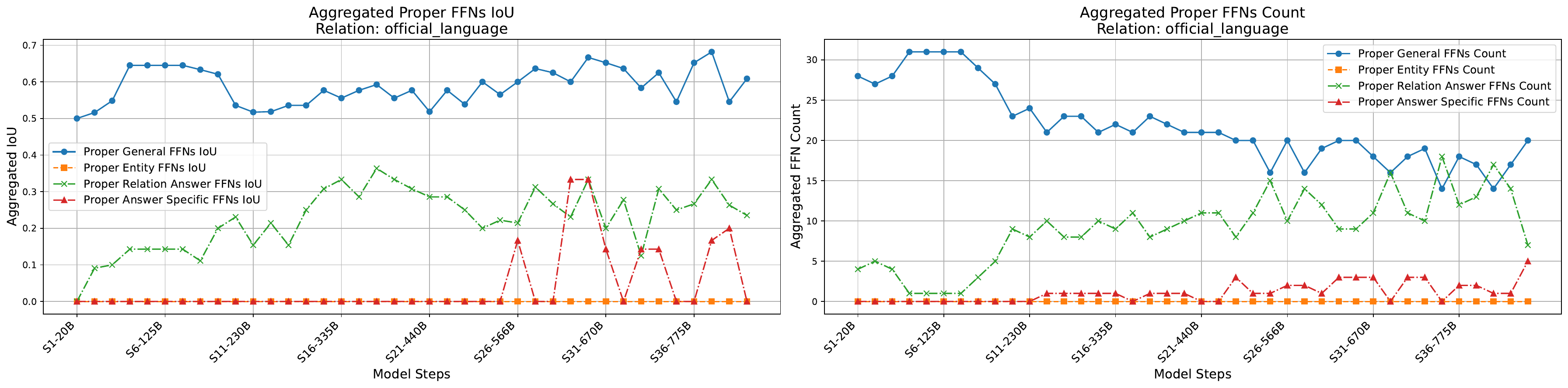}
    
    \begin{minipage}{\textwidth}  % Ensures full width
        \centering
        \caption{Relation-level FFN counts and IoU values.}
        \label{fig:all-subplotssecond}
    \end{minipage}

\end{figure}

\clearpage
\begin{figure}[h!]
    \centering
    \ContinuedFloat % Continues numbering from the previous figure
    \includegraphics[width=\textwidth]{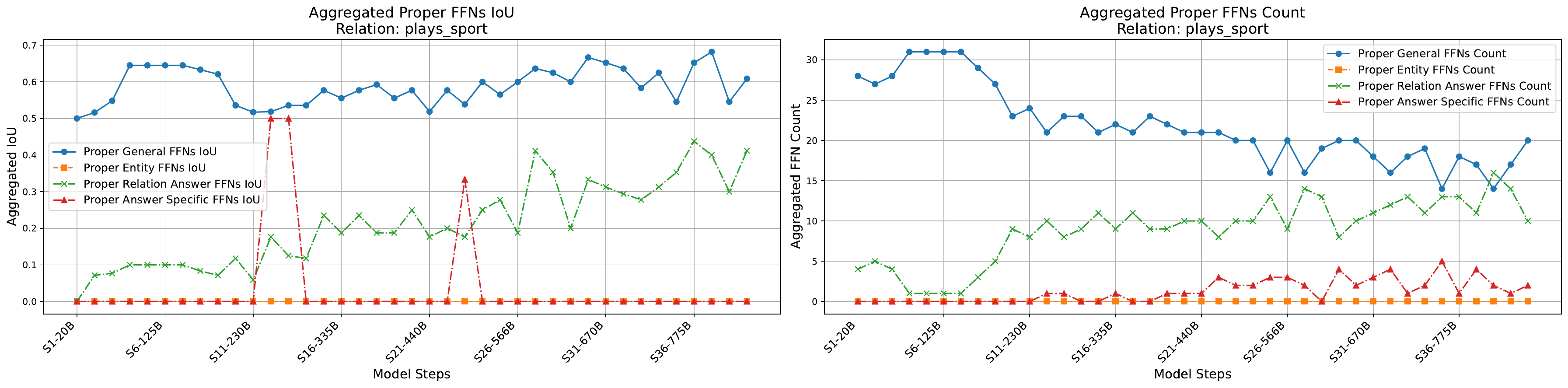}
    
    \vspace{0.3cm}
    \includegraphics[width=\textwidth]{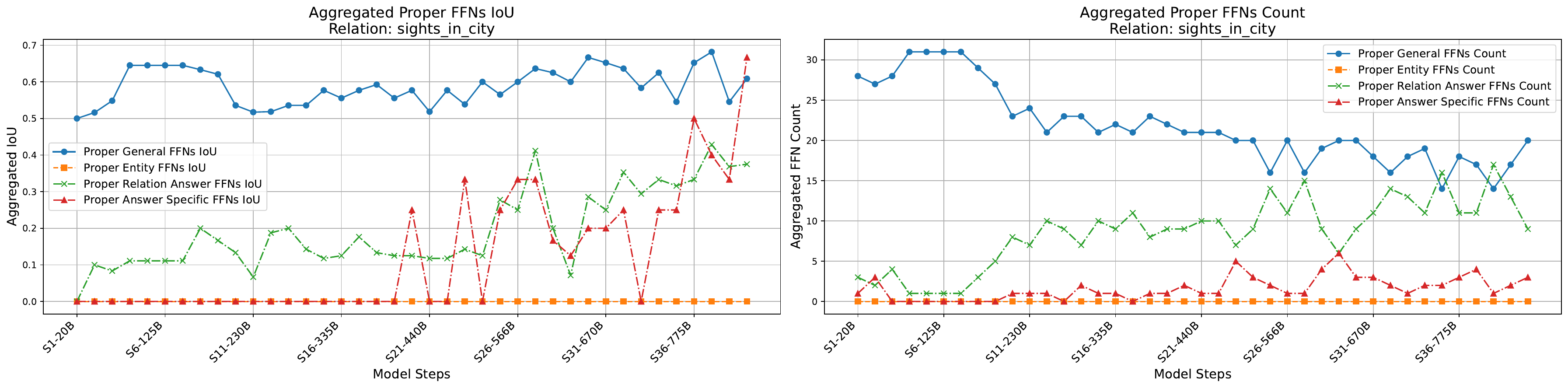}
    
    \vspace{0.3cm}
    \includegraphics[width=\textwidth]{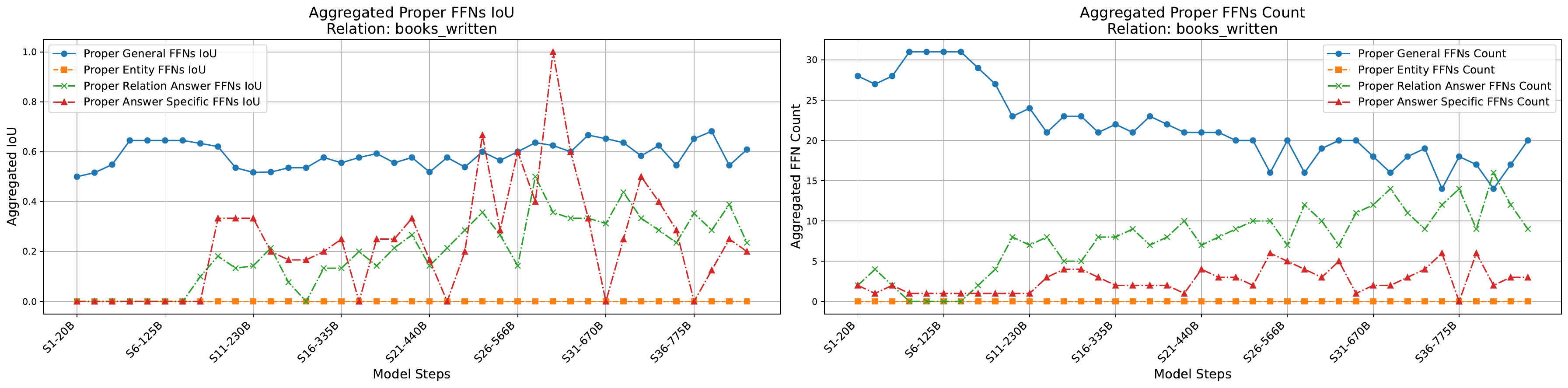}
    
    \vspace{0.3cm}
    \includegraphics[width=\textwidth]{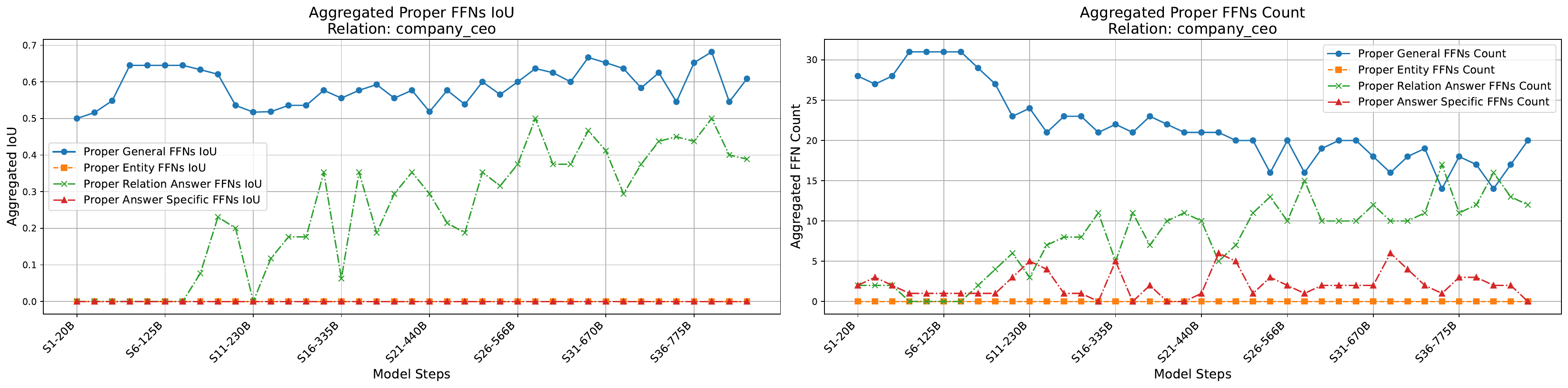}
    
    \vspace{0.3cm}
    \includegraphics[width=\textwidth]{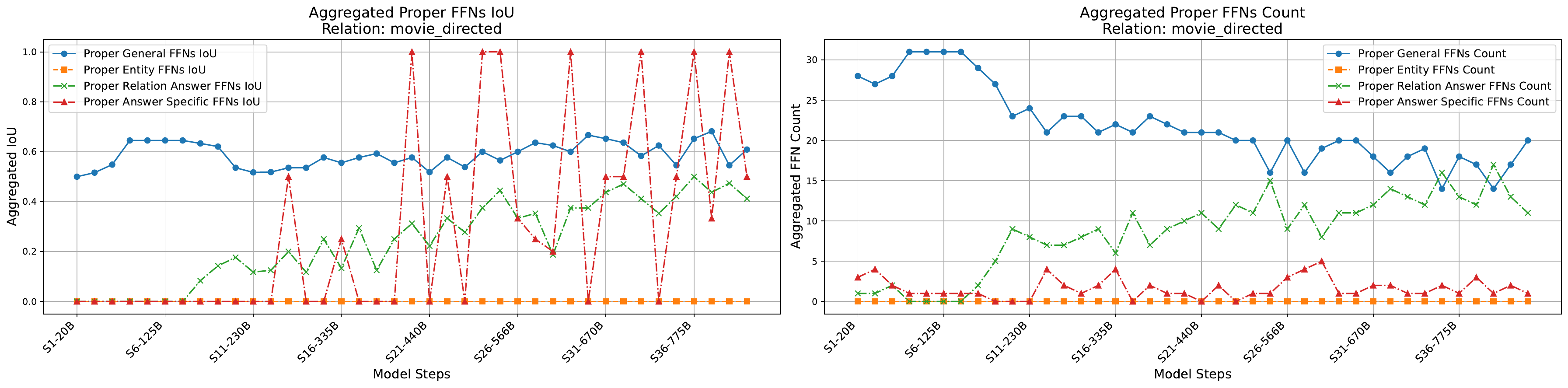}
    \begin{minipage}{\textwidth}  % Ensures full width
        \centering
        \caption{(continued) Relation-level FFN counts and IoU values.}
    \end{minipage}
\end{figure}

\clearpage

\section{Attention Head Switches}
\label{app:attention-head-switches}

\begin{figure}[h]
    \centering
    \includegraphics[width=\textwidth]{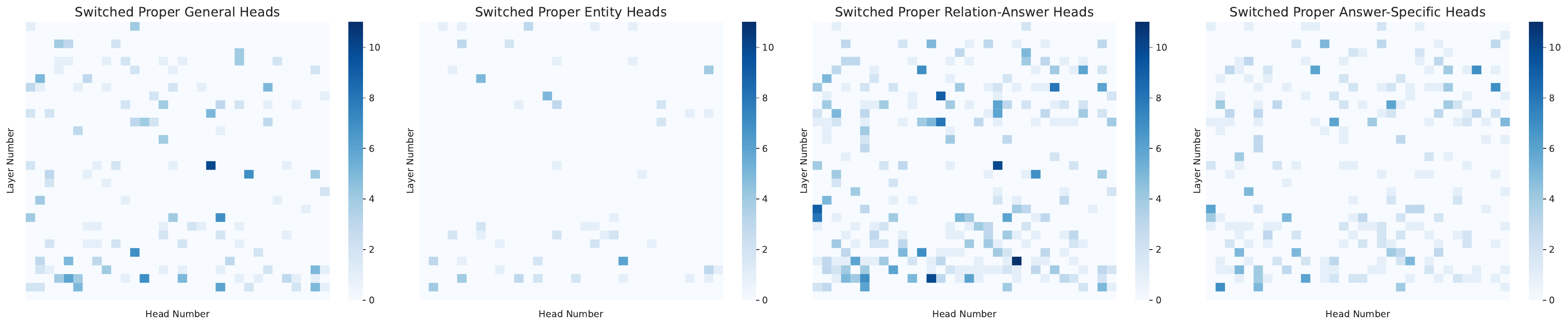}
    \begin{minipage}{\textwidth}  % Ensures full width
        \centering
        \caption{Accumulated head switches for LOC relations, independent of switch type.}
        \label{fig:switches-loc_app}
    \end{minipage}
\end{figure}

\begin{figure}[h]
    \centering
    \includegraphics[width=\textwidth]{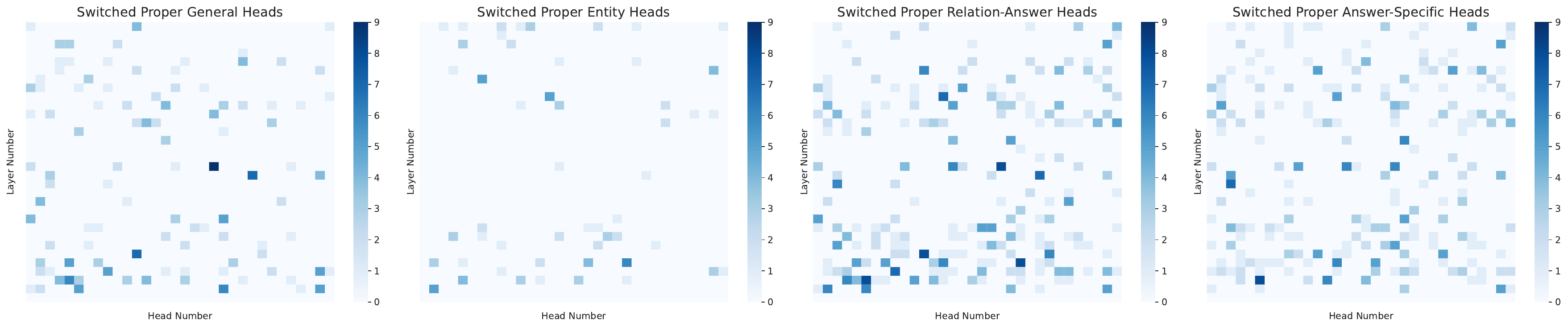}
    \begin{minipage}{\textwidth}  % Ensures full width
        \centering
        \caption{Accumulated head switches for NAME relations, independent of switch type.}
        \label{fig:switches-name_app}
    \end{minipage}
\end{figure}

In the following two plots, we observe that, as seen in the aggregated figure, layers 10–18 exhibit fewer switches, while switches occur more frequently in the early (0–10) and late (18–31) layers. However, when examining transitions between relation-answer and answer-specific roles, a clear distinction emerges: NAME-based relations (Fig.~\ref{fig:switches-name_app}) show significantly more switches in layers 10–18 compared to LOC-based relations~(\ref{fig:switches-loc_app}). Additionally, NAME-based tasks involve a greater number of distinct attention heads during these transitions. A switch refers to the reallocation of an attention head from one role to another among the four predefined roles.

\clearpage

\clearpage

\section{FFN Role Transition Count and Transition Probability}
\label{app:ffn-role-transitions}

\begin{figure}[h!]
    \centering
    \includegraphics[width=\columnwidth]{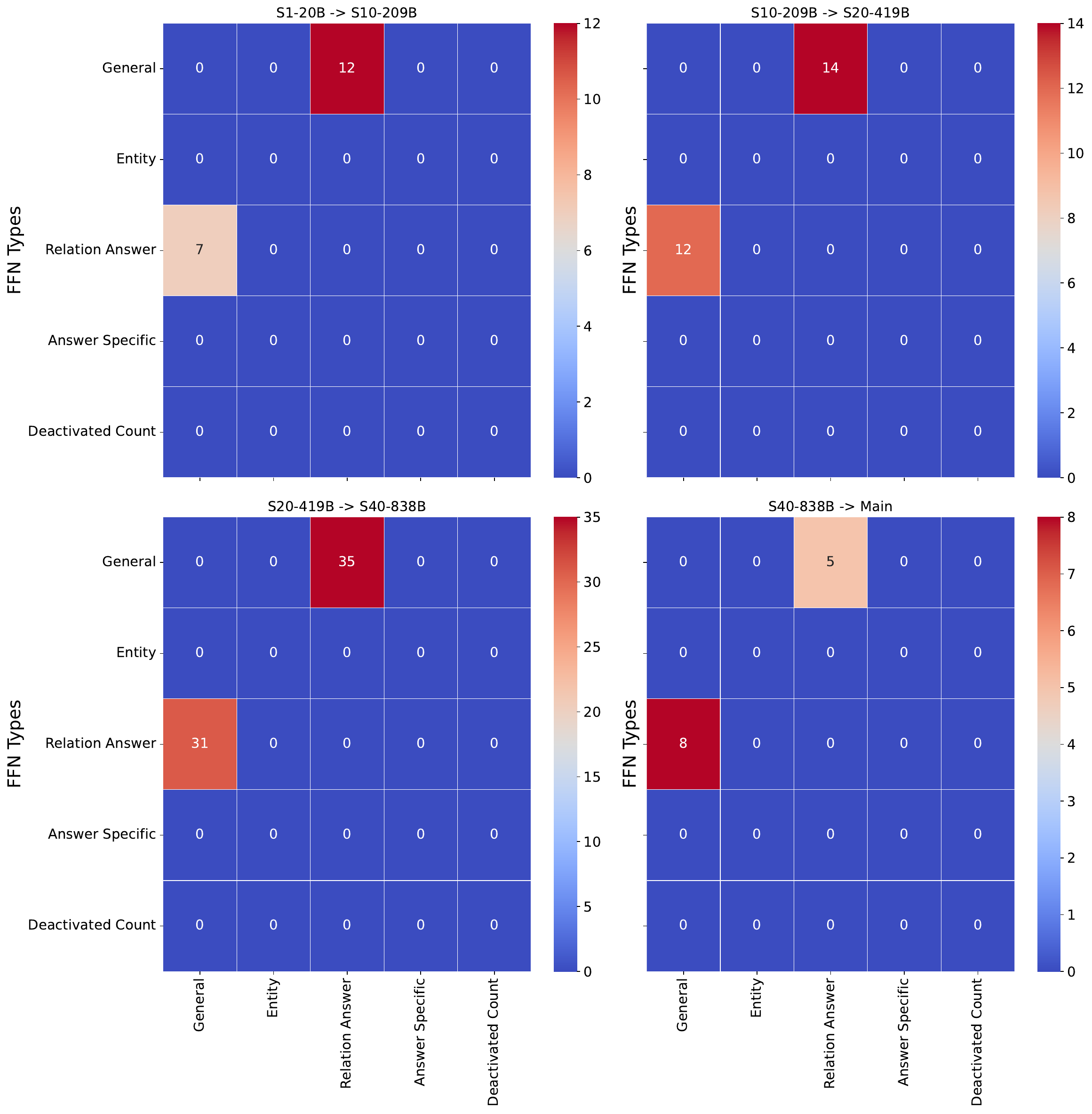}
    \caption{FFN Role Transitions. Heatmaps showing the frequency of role switches among proper general, entity, relation-answer, and answer-specific FFNs across layers.}
    \label{fig:accumulated_switch_counts_ffn_app}
\end{figure}

\begin{figure}[h!]
    \centering
    \includegraphics[width=\columnwidth]{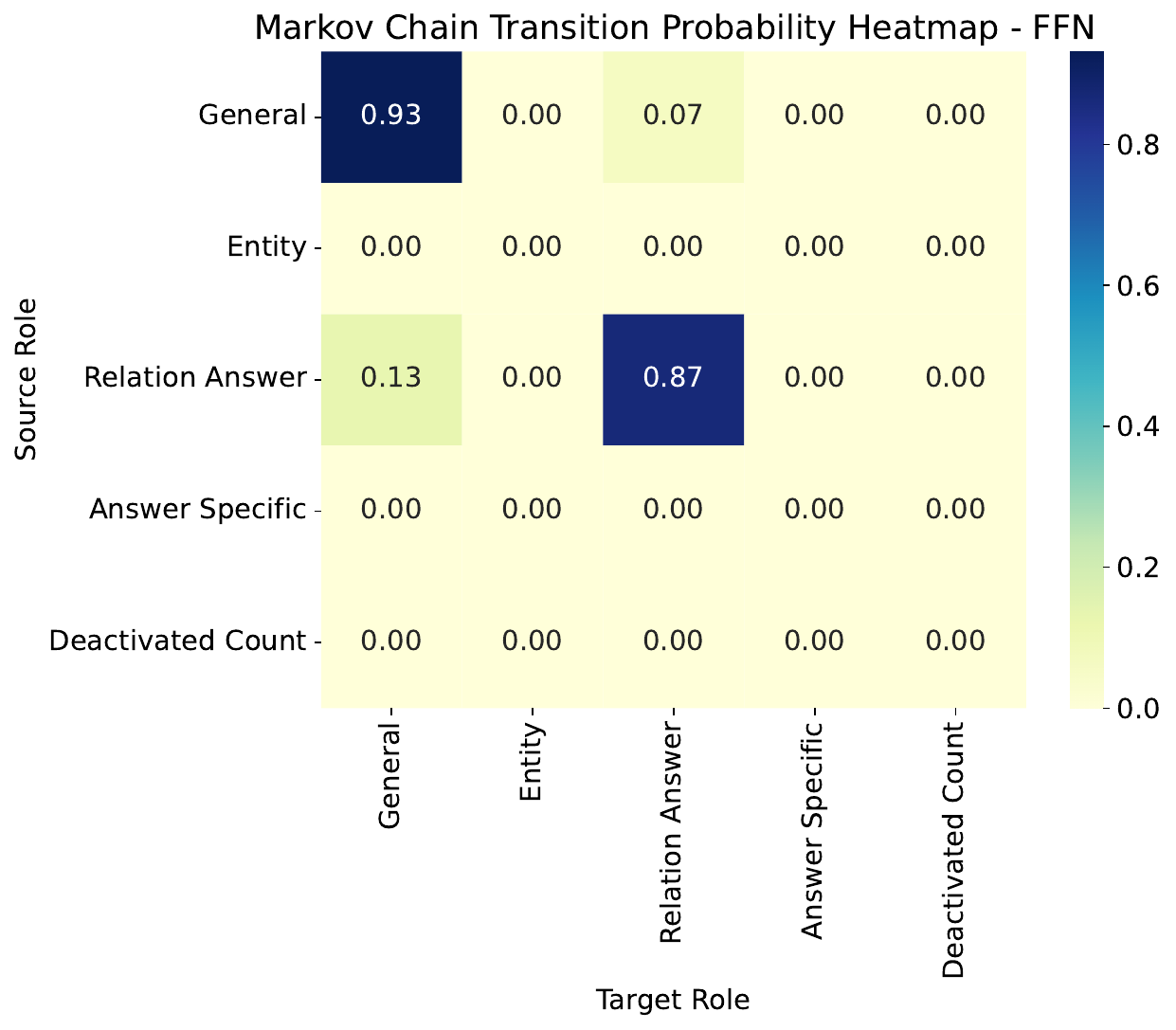}
    \caption{Markov Chain Transition Probability Heatmap showing the transition probabilities between different FFNs roles across model snapshots. Each cell represents the probability of a FFN transitioning from a source role (rows) to a target role (columns).}
    \label{fig:transition_probability_heatmap_all_app}
\end{figure}

\begin{figure}[h!]
    \centering
    \includegraphics[width=\columnwidth]{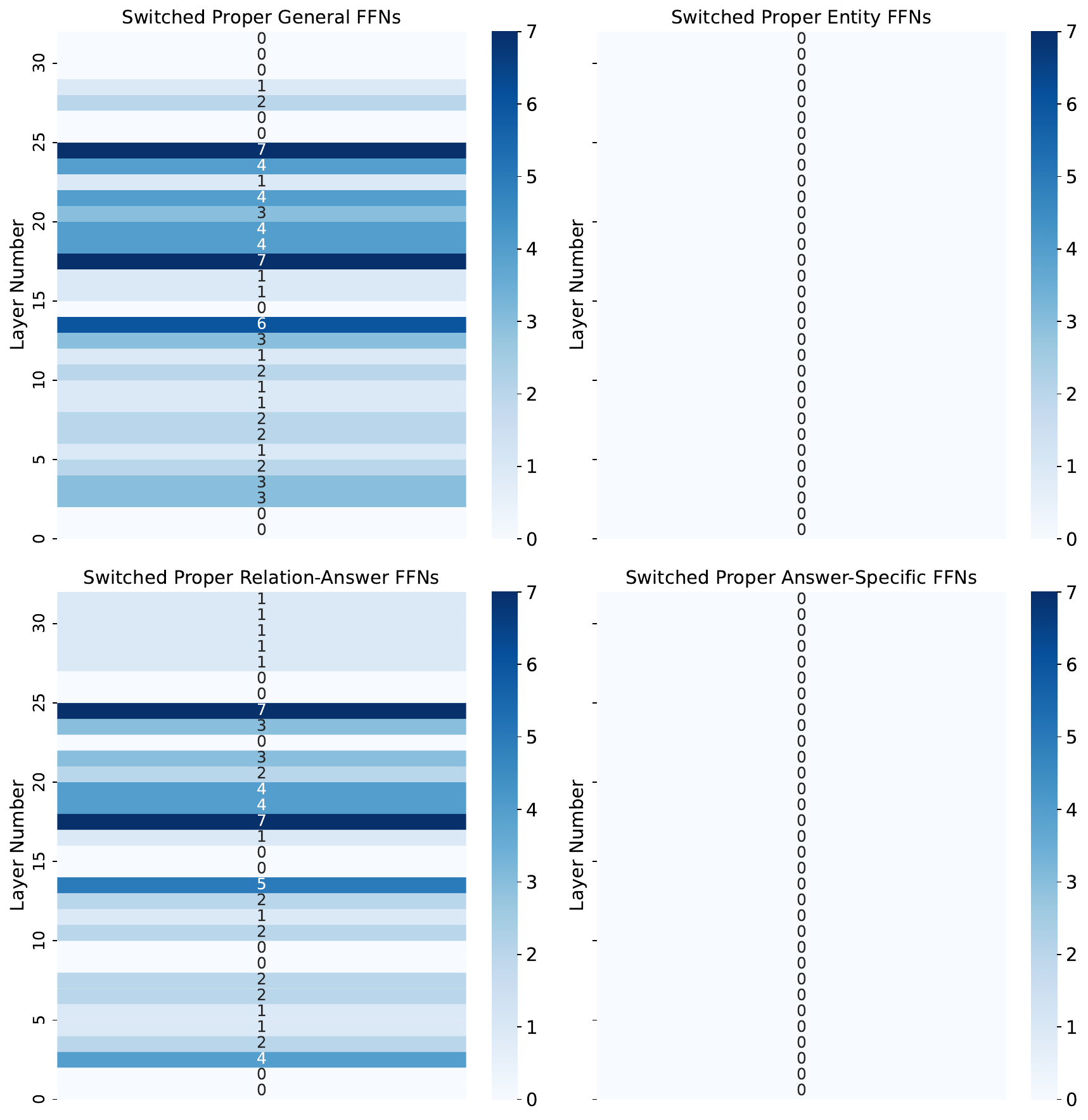}
    \caption{Layer-wise analysis of FFN role switching.}
    \label{fig:ffn_switches_aggregated_app}
\end{figure}

The following three figures present the metrics and methods used to assess component dynamics. As shown in Figure~\ref{fig:accumulated_switch_counts_ffn_app}, very few switches occur overall, with most transitions happening between general FFNs and relation-answer FFNs. Examining the transition probabilities (see Fig.~\ref{fig:transition_probability_heatmap_all_app}) from our Markov chain analysis, we find that both general and relation-answer FFNs tend to remain in their current roles with high probability; when switches do occur, they are predominantly between these two roles. Additionally, a layer-wise analysis (see Fig.~\ref{fig:ffn_switches_aggregated_app}) reveals a stark contrast with attention heads: starting from the middle layers onward, FFN role switches are nearly absent, and their roles become firmly established. Entity and answer-specific FFNs exhibit minimal switching across all layers.

\end{document}